\documentclass[preprint,12pt]{elsarticle}

\usepackage{amssymb}
\usepackage{amsmath}

\usepackage{natbib}
\usepackage{algorithm}
\usepackage{algpseudocode}
\usepackage{float}
\usepackage{calc,amsmath,amssymb,amsfonts}
\usepackage[LGR,T1]{fontenc}
\usepackage[greek,english]{babel}
\usepackage{xcolor,longfbox,fancyhdr}
\usepackage{array,supertabular,hhline,enumitem,hyperref}
\usepackage{cleveref}
\hypersetup{colorlinks=true,allcolors=blue}
\makeatletter
\newcommand\arraybslash{\let\\\@arraycr}
\makeatother

\journal{Artificial Intelligence}

\begin{document}

\begin{frontmatter}



\title{Unlock Reliable Skill Inference for Quadruped Adaptive Behavior by Skill Graph}


\author{Hongyin Zhang$^{a,b,*}$, 
Diyuan Shi$^{b,*}$, 
Zifeng Zhuang$^{b}$, 
Han Zhao$^{b}$, Zhenyu Wei$^{b,c}$, Feng Zhao$^{b}$, Sibo Gai$^{b}$, 
Shangke Lyu$^{b,**}$, 
Donglin Wang$^{b,**}$
} 

\cortext[cor1]{Equal contribution.}
\cortext[cor2]{Corresponding author. \{lyushangke, wangdonglin\}@westlake.edu.cn}

\affiliation[1]{organization={Zhejiang University},
city={Hangzhou},
country={China}}

\affiliation[2]{organization={School of Engineering, Westlake University},
city={Hangzhou},
country={China}}

\affiliation[3]{organization={Westlake Robotics},
city={Hangzhou},
country={China}}

\begin{abstract}
Developing robotic intelligent systems that can adapt quickly to unseen wild situations is one of the
critical challenges in pursuing autonomous robotics. Although some impressive progress has been made in walking
stability and skill learning in the field of legged robots, their ability for fast adaptation is still inferior to that
of animals in nature. Animals are born with a massive set of skills needed to survive, and can quickly acquire new
ones, by composing fundamental skills with limited experience. Inspired by this, we propose a novel framework, named
Robot Skill Graph (RSG) for organizing a massive set of fundamental skills of robots and dexterously reusing them for
fast adaptation. Bearing a structure similar to the Knowledge Graph (KG), RSG is composed of massive dynamic behavioral
skills instead of static knowledge in KG and enables discovering implicit relations that exist in between the learning
context and acquired skills of robots, serving as a starting point for understanding subtle patterns existing in
robots' skill learning. Extensive experimental results demonstrate that RSG can provide reliable
skill inference upon new tasks and environments, and enable quadruped robots to adapt to new scenarios and quickly
learn new skills.
\end{abstract}




\begin{keyword}


Robot Skill Graph \sep Massive Skills Organization and Query \sep Skills Inference
and Execution \sep New Skill Adaptation and Fast Learning
\end{keyword}

\end{frontmatter}

\section{Introduction}
Animals are born with various fundamental skills to survive in complex and varied wild environments, and can swiftly
learn new ones by composing these skills with little effort or experience. For instance, sea turtles are born on the
shore but may swiftly pick up the new ability to swim when they first enter the sea by adjusting the way their flippers
swing, while most legged animals are naturally able to gallop over various terrains without tripping. 
Animals never learn these extraordinary abilities from scratch for specific tasks. 
The mechanism with
which an animal can acquire extra sensorimotor abilities while adjusting to external environments implies that these
massive fundamental skills lay a solid foundation for fast learning new skills.

Recently, the main goal of Artificial Intelligence (AI) and robotics systems has been to achieve human-level
intelligence and behaviors. 
Despite recent breakthroughs of AI in decision-making~\cite{1_silver2016mastering,2_silver2018general,3_schrittwieser2020mastering,4_berner2019dota,5_vinyals2019grandmaster}, healthcare~\cite{6_yu2018artificial}, legged
locomotion~\cite{7_choi2023learning}, and content generation~\cite{8_ramesh2022hierarchical,9_ouyang2022training,10_saharia2022photorealistic} tasks, many of the basic sensorimotor capacities that animals
have or acquire effortlessly, remain deceptively challenging for robotic systems~\cite{11_cully2015robots}. This inferiority is partially
due to AI systems' fragility to unforeseen changes and lack of ability to interact with
unpredictable events~\cite{11_cully2015robots,12_carlson2005ugvs}. Therefore, to succeed in a world of unknowns, an agent must adapt to new situations using
acquired knowledge.

\begin{figure}[h]
\centering
\includegraphics[width=0.9\textwidth]{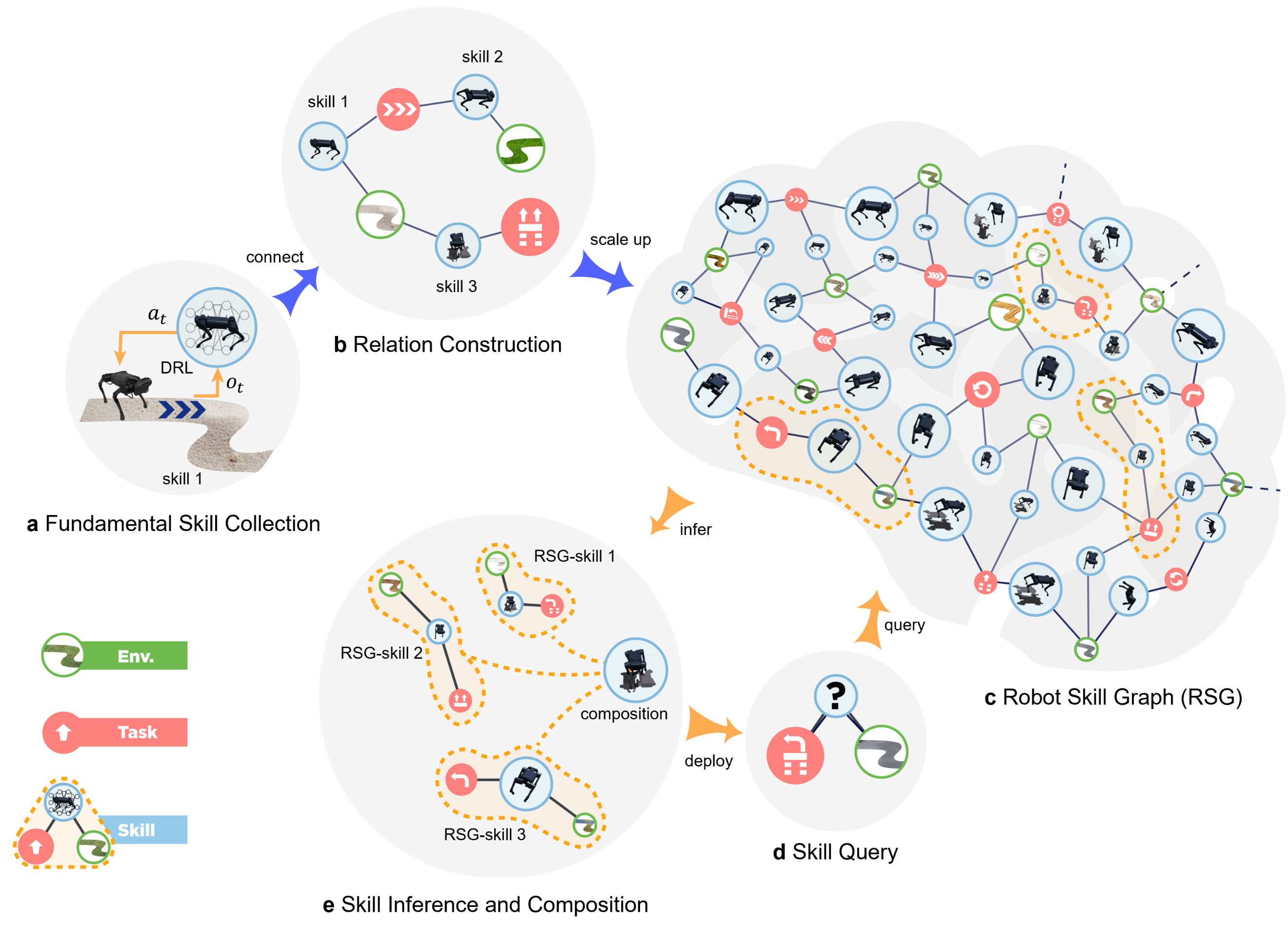}
\caption{\textbf{Overall framework for the construction and application of
RSG.} \textbf{a}, Collect a rich and diverse set of fundamental skills through the DRL approach with each skill
consisting of a task, an environment and a policy network. \textbf{b}, Different skills are connected through
relationships between environmental entities and task entities. \textbf{c}, Illustration of the RSG structure.
\textbf{d}, Given a new task and environment query, the RSG calculates the match between existing fundamental skills
and new required skills. \textbf{e}, Finally, these skills inferred by RSG will be executed, composited, or fine-tuned respectively according to the matching degree. The newly learned skills will be also added to RSG for future usage. }
\label{fig_Overall framework for the construction and application of RSG.}
\end{figure}

However, many existing quadruped robot strategies concentrate on learning from scratch or by transferring a few
disposable skills. In comparison, animals in nature have a large number of survival skills and can perform
skill-related reasoning and execution smoothly and efficiently. In this paper, we investigate developing quadruped
robots that can mimic the adaptive behavior of animals in nature by leveraging a variety of massive fundamental skills
(or behaviors) to adapt to varied environments and tasks. By drawing inspiration from the Knowledge Graph (KG) and our
previous original concept Skill Graph~\cite{13_zhao2022ksg}, we propose a new framework called Robot Skill Graph (RSG) for the
organization, query, inference, and execution of massive skills for quadruped robots, as well as the rapid learning of
novel skills
(Fig.~\ref{fig_Overall framework for the construction and application of RSG.}).

The RSG can efficiently organize massive fundamental skills that robots have already mastered. Given a new
environment or task, RSG queries it as input and infers the most appropriate skills as output. The fundamental reason
behind this approach is that existing skills in RSG contain plentiful prior knowledge that is valuable for new
scenarios, and thus can provide feasible initialization by correctly extracting and coordinating relevant skills. These
skills culled from the RSG will serve as a basis for further rapid learning to complete new tasks quickly. To sum up,
the proposed RSG-based motor skill generation mechanism is evaluated on the quadruped robot, which aims to pursue a
flexible motion behavior in front of unpredictable changes and meanwhile acquire fast and flexible adaptive motion
capabilities for new scenarios.
The core contributions of our work include five aspects:
\begin{itemize} 
\item \textbf{Wide-Coverage Skill Diversity}: A novel Robot Skill Graph (RSG) framework encodes 320 diverse quadruped skills across 12 environments and 31 tasks, greatly surpassing previous work limited to specific locomotion or environmental adaptability.
\item \textbf{Robust Skill Inference and Execution}: RSG enables reliable skill inference and execution in long-horizon sequence tasks, even under dynamic perturbations, such as rapid recovery from falls in parkour tasks.
\item \textbf{Autonomous Skill Discovery}: Visual-based environmental perception is organically integrated with RSG to enable efficient skill selection and autonomous decision-making capabilities for robots.
\item \textbf{Rapid Adaptation of Complex Behaviors}: Robust skill representation based on knowledge graph embedding combined with Bayesian optimization enables rapid adaptation of new skills in seconds.
\item \textbf{Efficient Learning of Unseen Skills}: Rapidly fine-tune unseen complex skills in minutes through online reinforcement learning, with sample efficiency far exceeding training from scratch.
\end{itemize}

\section{Related work}
\textbf{DRL-based Quadrupedal Locomotion:} In the Deep Reinforcement Learning (DRL)-based quadrupedal locomotion
research, prior knowledge is represented in a variety of forms, such as motion data~\cite{14_singla2019realizing,15_peng2020learning,16_vollenweider2023advanced}, trajectory generators~\cite{17_iscen2018policies,18_jain2019hierarchical,19_rahme2020dynamics,20_zhang2021terrain}
and control methods~\cite{21_yang2022fast,22_lyu2023composite,23_yao2021hierarchical}. \ In some recent important studies, researchers either
focused on achieving robust walking in diverse environments~\cite{24_lee2020learning,25_nahrendra2023dreamwaq} or explored skill training and execution for
specific tasks~\cite{26_yang2020multi,27_jin2022high,28_han2024lifelike}. These prior studies mainly consider DRL policy training and execution for specific tasks, while
our work covers a wider range of environmental terrains and tasks.

\textbf{Skill-based Robot Learning:} In skill-based DRL, skills are generally represented as sub-policies or a series of
low-level actions to facilitate learning long-horizon behaviors. Many works propose instructing agents to take action
as temporal-extending skills, such as options~\cite{29_sutton1999between,30_shankar2020learning,31_shankar2022translating} or motion primitives~\cite{32_pastor2009learning,33_pertsch2021accelerating,34_salter2022priors,35_rao2021learning,36_pertsch2021guided,37_nam2022skill,38_adeniji2022skill,39_rana2023residual}. However, skill-based RL struggles
with more challenging real-world tasks and requires a large number of environmental interactions~\cite{40_lee2021adversarial,41_huang2023skill}. \ Moreover, Previous work~\cite{11_cully2015robots} built a detailed map of the
high-performance behavior space containing a large number of skills (skills are determined only by subtle differences
between gaits) and designed an intelligent trial-and-error algorithm that allows the robot to adapt quickly. Later work~\cite{42_chatzilygeroudis2018reset} also built a skill set that allows the robot to recover from damage while completing a task, but their task is
only to reach a certain position in the 2D plane. Therefore, the range of skills considered in previous studies is
relatively limited, while our work shows a wider range of skill diversity to cope with various complex unstructured
scenarios.

\textbf{Knowledge Representation Learning:} Knowledge Representation Learning (KRL), a.k.a. Knowledge Graph Embedding
(KGE) \cite{43_ji2021survey,44_ali2021bringing} aims to learn latent representations of graph-structured data with \textit{entities} as nodes and
\textit{relations} as edges. As a classical model, TransE \cite{45_bordes2013translating} is proposed to learn data representations from
pre-defined inductive bias, then TransH \cite{46_wang2014knowledge} improved TransE by supporting one-to-many, many-to-one, and many-to-many
relations prevalent in large-scale graph data. In our work, we model \textit{environments}, \textit{tasks},
\textit{skills} and their \textit{relationship }as\textit{ }complex, graph-structured data and employ TransH-based KRL
for RSG representation learning.

\section{Preliminaries}
\textbf{Knowledge Representation Learning}. Let's consider a knowledge graph  $G=\{E,R,F\}$ where $E=\{e\}$ is the set of all entities, $R=\{r\}$ is the set of all relations and $F=\{f=(e_h,r,e_t)\}$ is a (possibly incomplete) set of facts $f$ with $e_h$ being a head entity and $e_t$ being a tail entity. The goal of Knowledge Representation Learning is to learn the representation of all $e$ and $r$ given $F$ and a score function $\mathbb{S} = (e_h, r, e_t) \rightarrow \mathbb{R}$, such that $\mathbb{S}(f)$ is a in a high score for any $f \in F$.

\textbf{TransE and TransH}. As a classical KRL method, TransE  defined the score function as $\mathbb{S}_{\text{TransE}}=-\vert\vert e_h +r -e_t\vert\vert_2^2$. Then TransH \cite{46_wang2014knowledge}  improved over TransE with the ability to
model one-to-many and many-to-one relations by performing translation in a relation-dependent hyperplane: $\mathbb{S}_{\text{TransH}}=-\vert\vert (e_h-w^T_re_hw_r) + d_r - (e_t-w_r^Te_tw_r)\vert\vert_2^2$ where $w_r$ is the norm of hyperplane and $d_r$ is a translation vector of it.

\textbf{Knowledge Graph Completion}. Given a trained $G$, Knowledge Graph Completion (KGC) focuses on
discovering “missing” facts of it: for any triplet $(e_h, r, e_t)$ which may be out of $F$, a high score $\mathbb{S}(e_h, r, e_t)$ indicates this triplet is likely to be true  (the $e_h$ has a $r$ relation to $e_t$) while a low score indicates false triplet.

\section{Method}
The construction and application of RSG can be divided into three main stages (Fig.~\ref{fig_Overview of the presented RSG.}). First, we describe
the method of collecting a large number of fundamental skills, including locomotion skill types and more flexible skill
types. Then, based on a variant of the KRL model, we characterize how to establish the relationship between skills,
environment, and task entities and form a skill graph. Finally, we show how to calculate skill inference scores and
induce different skill application modes.

\subsection{Fundamental Skill Collection}
\label{method:Fundamental Skill Collection}
\begin{figure}[H]
\centering
\includegraphics[width=0.78\textwidth]{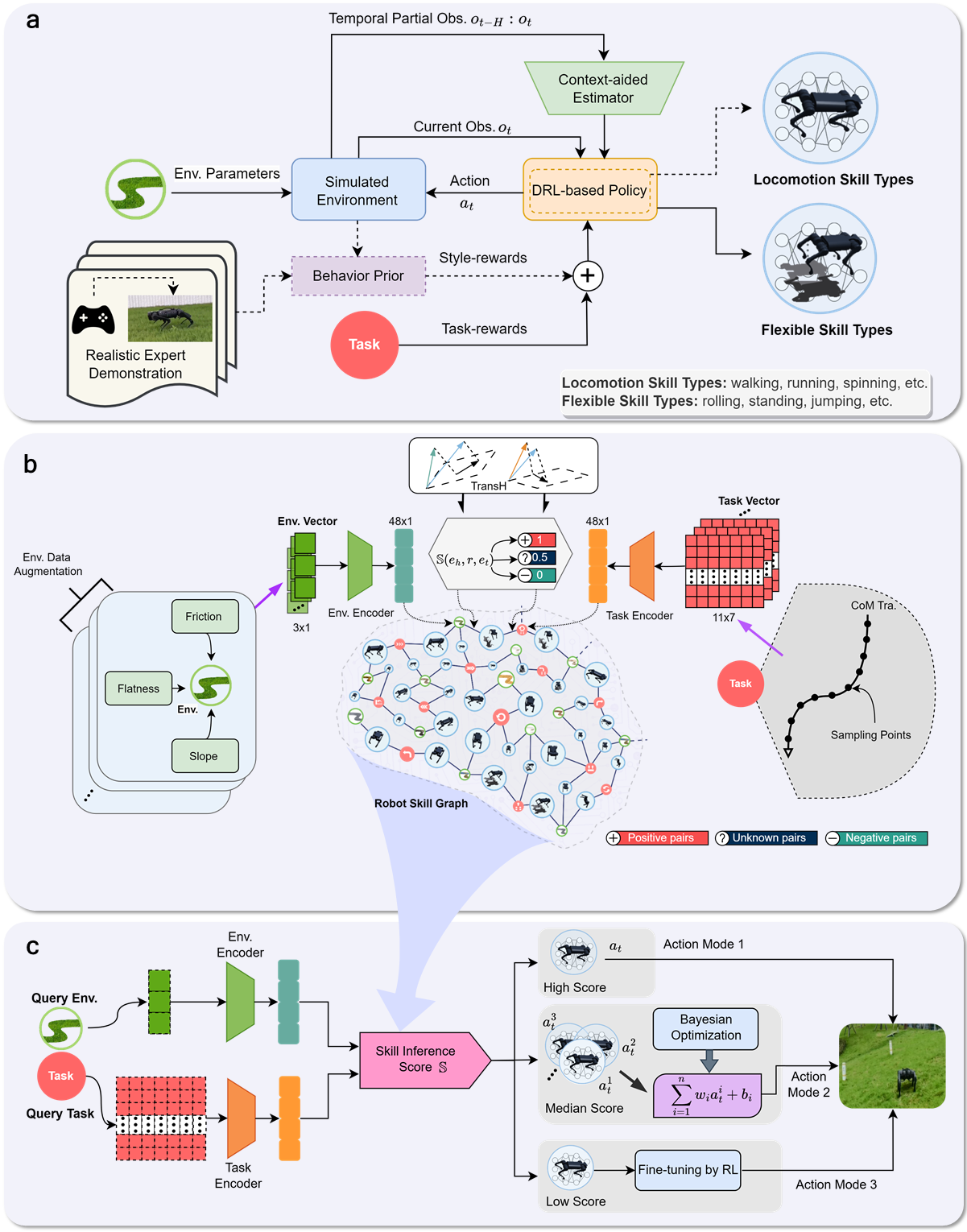}
\caption{\textbf{Overview of the presented RSG.} \textbf{a,} Fundamental skills are divided into locomotion
skill types (walking, running, etc.) and flexible skill types (rolling, posture recovery, etc.). The context-aided
estimator network (CEN)~\cite{25_nahrendra2023dreamwaq} is used to enhance the environmental representation. We also utilize adversarial motion
priors (AMP)~\cite{50_escontrela2022adversarial} to provide expert-style behavior for locomotion skill tasks. \textbf{b,} To construct the RSG, the
environment is described as friction, flatness, and slope, and the task is described as eleven consecutive CoM
trajectory points. The environment and task descriptions are then mapped into latent variables respectively (with 48
dim. here as a hyperparameter). The TransH is leveraged to complete the relationship construction between skill
entities, task entities, and environment entities. 
\textbf{c,} For a new skill query, we use a score function to measure the match between the required new skill and the existing fundamental skills (i.e., skill inference score).
For the application, according to the high, medium, and low scores of the inferred skills, the action modes adopt the
methods of skill execution, BO composition, and RL finetuning respectively.
}
\label{fig_Overview of the presented RSG.}
\end{figure}

To make the skill
distribution sufficiently rich and diverse, we model the robot's behavior in the specific environment and task as a Markov decision process and utilize DRL-based methods to obtain skill
policies (Fig.~\ref{fig_Overview of the presented RSG.}\textbf{a}). 
Environment and task design first need to be considered.

The environmental design of legged robots mainly needs to consider terrain traversability~\cite{51_haddeler2022traversability}. Intuitively, from a
discrete perspective, the terrain can be assigned to separate categories, with each category having assumed mechanical
properties of the terrain. An instance of terrain classification may be the distinction of terrain traversability~\cite{52_bradley2015scene}.
We can also describe some terrain properties, assigning continuous values to the terrain, such as slope~\cite{53_stelzer2012stereo}, step
height~\cite{54_homberger2017terrain,55_wermelinger2016navigation}, or roughness~\cite{56_krusi2017driving,57_belter2019employing}. For continuous measurements, the terrain traversability can be based on
thresholding the values. For example, terrain traversability can be determined by individually thresholding terrain
slope, roughness, and step height~\cite{53_stelzer2012stereo}. Therefore, when building the simulated environment, we mainly consider the
roughness, bumpiness, and steepness of the realistic terrain. All terrains are defined in IssacGym~\cite{58_makoviychuk2021isaac} by three contact
properties: friction, flatness, and slope.
Environment parameters are in Appendix Tab~\ref{app:Environment parameters.}.

For task design, we consider common locomotion tasks and the more flexible behaviors of robots in the initial version of
RSG. Locomotion tasks include relatively periodic horizontal movements on the horizontal plane, such as walking,
running, spinning, etc. The main feature of this type of task is that the robot's body remains
basically parallel to the ground during its movement. More flexible behaviors are not limited to horizontal movements,
such as jumping, rolling, posture recovery, etc. Its main characteristics are high dynamics, flexibility, and agility.
In fact, any robot CoM trajectory can be used as a task type of RSG as long as the trajectory is available to the
robot. In this way, during the deployment phase, as long as we specify a feasible CoM trajectory at will, the robot can
immediately complete the corresponding complex behavior. More agile and dynamic, animal-like behavior will be built
into RSG in the future.


For locomotion tasks, we first collect some expert demonstrations in reality by utilizing an off-the-shelf remote
control to send body velocity commands. These expert demonstrations could provide diverse periodic motion behaviors
which is essential for developing skills with high quality. We adopt Inverse Reinforcement Learning (IRL) to convert
expert demonstrations into reward functions. Inspired by the AMP method~\cite{50_escontrela2022adversarial}, a discriminator network is also utilized to
distinguish expert demonstrations from policy-generated trajectories. AMP enables the policy to merge the acquired
expert signal with the task goal. The trained discriminator can be utilized as an intrinsic reward function for skill
policy training. The intrinsic reward function can guide the robot to learn a behavior style that is as close to the
real expert demonstration as possible. Therefore, the total reward function is expressed: 
\begin{equation}
r_t=w^{g}r_{t}^{g}+w^{s}r_{t}^{s},
\end{equation}
where $r_{t}^{g}$ is the task-specific reward and $r_{t}^{s}$ is the intrinsic reward generated by the AMP. $w^{g}$ and $w^{s}$
 are hyperparameters. For more flexible behaviors, the reward function is defined manually without utilizing AMP.
Moreover, the rolling and posture recovery behaviors themselves are highly dynamic and intense. Simply relying on
designing reward functions to smooth these behaviors is indirect and tricky~\cite{26_yang2020multi}. Therefore, we impose a penalty in the
DRL training loss function to encourage gentle behavior $J=J_{DRL}+\hat{J}$
 . The additional regularization term $\hat{J}$
 is:
\begin{equation}
\hat{J}=\left\|\bar{q}-\mu\left(s_t\right)\right\|,
\end{equation}
where the $\bar{q}$ is the given joint position and the $\mu\left(s_t\right)$
 is the action of the skill policy (i.e. the desired joint position). The definitions of all tasks and related reward
functions are in Appendix Tab~\ref{app:Name and definition of reward functions.} and Tab~\ref{app:Definition of tasks.}.

We utilize the classic model-free DRL algorithm, PPO~\cite{49_schulman2017proximal}, to train the skill policy. For the implementation of PPO, an
asymmetric actor-critic architecture is adopted~\cite{59_rudin2022learning}. That is, the actor network only receives observations that are
available in reality, while the critic network not only receives these observations but also receives privileged
information that is difficult to measure in the real world. Besides, a CEN~\cite{25_nahrendra2023dreamwaq} is utilized to strengthen environmental
representations The CEN is inserted into the PPO training, which encodes the observation history over the past
measurements to estimate the body velocity and implicitly infer the environmental representation. The CEN utilizes the
historical data in the replay buffer of PPO for supervised training, and its labels are the linear velocity of the
current time step and the observation of the next time step. More DRL training details are in   \ref{app:section_Data collection details, observations, and reward functions}.
Hyperparameters are in Appendix Tab~\ref{app:Definition of robot parameters and domain randomization.} and Tab~\ref{app:Hyperparameters of PPO algorithm.}.

\subsection{RSG Construction}
\label{method:RSG Construction}
Our RSG is constructed based on the framework of the KG. Normally, the KG consists of three components: \textit{entities}, \textit{relations}, and \textit{facts}. In RSG,  \textit{entities} include \textit{skills} ($s$), \textit{environments} ($e$), and \textit{tasks} ($t$) (In the following text, we use “context” ($c$) to denote both environment and task for brevity). \textit{Relations} considered in our setting are \textit{environment to skill} ($r_{e\rightarrow s}$) and \textit{task to skill} ($r_{t\rightarrow s}$). \textit{Facts} are  directly extracted from fundamental skill collection phase where a pre-trained skill is defined to have links to its
training environment and task. 
A distinctive feature of our build process is the construction of “class” information for context nodes. The
environments are represented by three properties: friction, flatness, and slope. We first define environment classes,
where the properties of each environment class are not fixed values but a range. This is because when the skill policy
is trained in simulation, we adopt domain randomization: a skill is trained in a bunch of similar
environments with slightly varying physical properties. Moreover, environmental description like “\textit{grassland}”
is intuitive but inherently vague. It could refer to various types of grassland with different levels of friction and
flatness. Hence to boost the robustness and generalability of RSG, for a given skill (e.g., “\textit{Forward Walking on
Grassland}”) we sampled 100 environment instances whose physical properties are concrete values from corresponding
environment class (“\textit{Grassland}”) and link these environment instances to the given skill. 

Similarly, to obtain an environment-agnostic task representation for task nodes, we rollout the trained skill policy 100
times in an anchor environment (\textit{indoor floor}). Here, the task classes are the probabilistic distributions of
the rollouts which are generally unknown and the 100 times of rollouts are task instances sampled from these underlying
distributions. All these 100 sampled rollouts are constructed as 100 task nodes and linked to given skill similar to
environment nodes. For elements of each task vector, we extract 11 timesteps in CoM's motion trajectory where each
timestep is a vector of 7 features. This essentially forms a profile of
robot's CoM velocity:
\begin{equation}
    \vec{t} = \biggl\{ \vec{t_i}=\Bigl[v_x^c , x_y^c, v_z^c, \vert \vert v^c \vert\vert, \mathbb{I}(\omega \ge 0), \mathbb{I}(\omega < 0), \vert \vert \omega \vert \vert\Bigr] \biggr\}_{i=1}^{11},
\end{equation}
where  $\vec{t}$ is the final task vector,  $\vec{t_i}$ is task vector in $i^{th}$ timestep, $[\cdots]$ forms a vector,   $v_x^c$, $v_y^c$ and $v_z^c$ are the normalized linear velocity components of the COM, $\vert \vert v^c \vert \vert$ is the norm of the linear velocity, $\mathbb{I}$ is the indicator function, $\mathbb{I}(\omega \ge 0)$, $\mathbb{I}(\omega <0 )$ form a one-hot encoding representing the direction of yaw velocity, and  $\vert\vert \omega \vert\vert$ is the norm of the yaw angle velocity. All 11 timesteps $\vec{t_i}$ are flattened to form the task vector $\vec{t}$ in $\mathbb{R}^{11 \times 7}$. 

\ \ In terms of skill nodes and relations, similar to the KRL, they are first initialized as random vectors and then
iteratively optimized against objective function. Additionally, we build environment and task encoders as multilayer
perceptrons. They are used to encode environment and task nodes into the same representation space as skills and
relationships and enable us to use prior, physical-based knowledge such as friction, and task trajectory along with
deep-learning-based representations.

\subsection{RSG Training}
\label{method:RSG Training}
As shown in Fig.~\ref{fig_Overview of the presented RSG.}\textbf{b}, to train our RSG, we first generate \textit{positive,} \textit{negative,} and
\textit{soft} triples as input data. Positive triples are indeed facts set $F$ in RSG. We then construct negative triples by constructing wrong-form triples, such as $(e, r_{e\rightarrow s}, e^\prime)$, $(t, r_{t\rightarrow s}, t^\prime)$,  $(e, r_{t\rightarrow s}, s)$ and $(t, r_{e\rightarrow s}, s)$. We also construct soft triples, as triples with various plausibility. This is achieved by flipping the head entity in
each positive triplet, such as $(e_{new}, r_{e\rightarrow s}, s_{original})$ and $(t_{new}, r_{t\rightarrow s}, s_{orignal})$. Then, by sampling from these triplets and feeding into RSG, we perform training with a contrastive loss function:
\begin{equation}
\mathcal{L} = \bigl(\mathbb{S}_{positive} -1)^2 + (\mathbb{S}_{negative} - 0 )^2 + \max(0, \mathbb{S}_{soft} -1 + \delta)\bigr),
\end{equation}
where $\mathbb{S}=e^{-\lambda \vert \vert (e_h - w_r^Te_hw_r)+d_r-(e_t-w_r^Te_tw_r) \vert\vert}$ is TransH's score function that could represent one-to-many and many-to-one relationships via
the projection of the entities on each relation-specific hyperplane (Fig. 1\textbf{b}), $\lambda=3$ is a hyperparameter, $\delta$ is proportional to the similarity $\kappa(c_n, c_o)$ between original context $c_o$ and the new context $c_n$:
\begin{equation}
    \delta \propto \kappa(e_o, e_n) = \max\bigl[ \frac{\vert \theta_o -\theta_n\vert}{2}, \text{norm}(\vert\vert f_o-f_n, \mu_o- \mu_n \vert\vert) \bigr]   \text{, and}
\end{equation}
\begin{equation}
    \delta \propto \kappa(t_o, t_n) = \max\Biggl\{ \sum_t \vert \text{sign}(\omega_{n,t}) -\text{sign}(\omega_{o,t}) \vert, \sum_t\biggl[1- \Bigl(\frac{v_{n,t}}{\vert v_{n,t}\vert}\Bigr)^T\Bigl(\frac{v_{o,t}}{\vert v_{o,t} \vert}\Bigr)\biggr] \Biggr\}, 
\end{equation}
where $\mu_i, f_i, \theta_i$ are friction, flatness, and slope of $i^{th}$ environment, $\text{norm}(x) = x / x_{\text{max}}$ normalizes a vector, $\text{sign}(\cdot)$ is the signature function, $\vert \cdot \vert$ is the absolute function, $\vert\vert ... \vert \vert$ represents a vector, and $\omega_{i,t}, v_{i,t}$ are angular velocity and linear velocity at timestep $t$ of $i^{th}$ task, respectively. Readers can refer to the original paper \cite{46_wang2014knowledge} for other auxiliary losses in the TransH model. 

A key design consideration of RSG is to enjoy the merits of both similarity-based methods like SBM and representation
learning methods like KRL: \textbf{(1) }Purely similarity-based methods could take advantage of interpretability and
prior domain knowledge, but struggle to model complex entity-relation dynamics and skill representation. \textbf{(2)}
On the contrary, vanilla KRL methods fail to utilize domain knowledge as all representation vectors are randomly
initialized and learned purely via optimization against objective function. This results in a black-box model that
lacks interpretability and doesn't generalize well to unseen scenarios. Indeed, vanilla KRL
methods often fall short in generalizing to new nodes (a.k.a. Out-of-Knowledge-Graph (OOKG) problem \cite{60_wang2023iteratively,61_chen2023generalizing}) which
however is crucial and prevalent in our setting.

Therefore, by designing our RSG's objective function:
\begin{equation}
    \mathcal{L} = \underbrace{(\mathbb{S}_{positive} -1 )^2 + (\mathbb{S}_{negative} -0)^2}_{\text{Term I}} + \underbrace{\max(0, \mathbb{S}_{soft}- 1 +\delta))}_{\text{Term II}},
\end{equation}
we could perform standard KRL training thanks to “Term I” and also incorporate domain knowledge via $\delta$ in “Term II”. RSG construction and representation can be found in the Appendix Algo.~\ref{app:RSG construction and representation.}.

\subsection{Skill Inference and Composition}
\label{method:Skill Inference and Composition}
Skill inference tries to obtain the skills that are most likely to complete the given tasks and environments (Fig.~\ref{fig_Overview of the presented RSG.}\textbf{c)}. As an extension to KGC, at the inference time, the query environment and task vectors are first converted
into representations with individual encoder. Then, the trained score function calculates the score between required
skills and query entities. Effectively, we calculate $\mathbb{S}_{\left(t, r_{t \rightarrow s}, s\right)}^{\text {task }} $ and $\mathbb{S}_{\left(e, r_{e \rightarrow s}, s\right)}^{env.} $separately for all skills and multiply them together to obtain the final skill inference score:
\begin{equation}
\mathbb{S}=\mathbb{S}_{\left(t, r_{t \rightarrow s}, s\right)}^{\text {task }} \cdot \mathbb{S}_{\left(e, r_{e \rightarrow s}, s\right)}^{env.} \in[0,1].
\end{equation}
This score describes the degree of match between the required skills and the fundamental skills in RSG. When facing downstream tasks, RSG returns the candidate fundamental skills in descending order of scores. The asymptotic
time complexity of skill inference \ is  $\mathbb{O}(cN)$
 , where   $c$
 denotes the number of contexts considered and $N$ is the number of fundamental skills. Three skill application strategies are adopted according to skill inference scores.

When the score $\mathbb{S}$ is high ($\alpha_{high}=0.9\leq\mathbb{S}\leq1$)
 , we directly utilize the first skill inferred by RSG to solve downstream tasks. This is because entities similar to
the query environments and tasks have already appeared in the RSG. The skill policies corresponding to fundamental
skills can be generalized to required tasks. 

When the score $\mathbb{S}$  is median ($\alpha_{low}=0.7\leq\mathbb{S}<\alpha_{high}=0.9$)
 , the first few inferred skills are combined into a parameterized linear model in the action space and then optimized
by the BO method. In previous work~\cite{11_cully2015robots}, an intelligent trial-and-error algorithm based on BO optimization was studied
that allowed the robot to adapt to damage in a large search space within two minutes without self-diagnosis or
pre-specified contingency plans. Inspired by this work, we consider utilizing BO composition to achieve real-time
online adaptation of real robots. Moreover, their task is limited to gait adjustment when walking on flat ground and
cannot cope with more unstructured scenarios, while our work considers more diverse terrain environments and locomotion
tasks, thus covering a wider skill distribution. Assuming that the RSG in the skills inference phase has selected  
$n$ skills, the linear combination can be expressed as:
\begin{equation}
a_t^*=\sum_{i=1}^n w_i a_t^i+b,
\end{equation}
where the $a_t^i$ is the action from the $i^{th}$ fundamental skills. The weight  $w_i$ and bias $b$ are the parameters of this linear model, and   $\sum_{i=1}^n w_i=1$. The score of selected fundamental skills by RSG inference is utilized as the initial value of weight optimization.
All the parameters are optimized by BO so as to balance training speed and effectiveness. Essentially,
the main function of the BO weight is to composite fundamental skills and establish links between
foundation skills and newly learned skills. The main function of BO bias is to provide the ability to learn new skills
to adapt quickly to new tasks (or environments). Furthermore, the reason why the BO method can be optimized in real
time is that it performs the linear transformation on the action space of the skill policy corresponding to the
fundamental skills. This makes the action space of the skill more flexible, allowing it to adapt to the environment or
tasks more quickly. 
More details are in   ~\ref{app:section_Skill composition by using Bayesian optimization}.

When the score $\mathbb{S}$  is low ($0\leq\mathbb{S}<\alpha_{low}=0.7$)
 , the first few inferred skills will be finetuned by the RL (i.e., PPO) rather than learning from scratch. This
situation arises because the query environment or task is out of distribution compared to the existing fundamental
skills in RSG. Specifically, the action of the new skill can be expressed as the linear weighted sum of the actions of
each fundamental skill: 
\begin{equation}
a_t^*=\sum_{i=1}^n w_i \pi_{\varphi_i}\left(\cdot \mid s_{i, t}\right)+b,
\end{equation}
where parameters $w_i$
 ,  $b$
 , and $\varphi_i$ all need to be updated by RL fine-tuning. $ \pi_{\varphi_i}\left(\cdot \mid s_{i, t}\right)$ represents the skill policy corresponding to the  $i_{th}$
 skill inferred by RSG. The value network of the fundamental skill with the highest score is fine-tuned as the initial
value. The new skill obtained during the skill composition stage will be added to the RSG with its environment and task
for RSG evolution and continuous learning. The expansion mechanism simulates the evolution in nature and the RSG will
consist of more and more skills, environments, and tasks as usage increases. The RSG inference and usage process is in
Appendix Algo.~\ref{app:RSG inference and usage.}, and hyperparameters are in Appendix Tab~\ref{app:Hyperparameters of the RSG inference and composition.}. 

The BO composition and RL finetuning methods utilize the same reward function form. The desired values
\hspace{0pt}\hspace{0pt}corresponding to different new skills are different, which depends on the 11 CoM trajectory
points of the new skill to be learned. The optimization objectives include task response terms, posture stability
terms, action smoothness terms, etc. 
Details are in  ~\ref{app:section_The optimization target for BO composition and PPO finetuning}. 
The alleviation of the sim2real problem
is summarized in  ~\ref{app:section_The alleviation of the sim2real problem}. 

In terms of task queries for new skills required for downstream tasks, RSG flexibly supports several specification
methods: manual calculation and automatic calculation by drawing trajectory sketches. For manual calculation, we can
calculate based on prior knowledge. For automatic calculation by trajectory sketches, inspired by previous research~\cite{62_gu2023rt}, we can manually draw the motion trajectories required by CoM with a mouse, and then convert these trajectories
into the required task queries. On the other hand, for environmental queries of required new skills, we can acquire
them manually through prior knowledge or autonomously utilizing multimodal LLMs (e.g., Gemini-1.5-pro~\cite{48_team2024gemini}). With the rapid
development of LLM in recent years, we can utilize it as an external visual perception and understanding module. By
setting reasonable prompts and utilizing the images of the first-person real-time camera on the robot to identify the
environmental terrain, environmental queries can be obtained in real time to achieve autonomous selection and inference
of skills.

\section{Experiments}
Enhancing accurate responsiveness to diverse unstructured environments and rapid adaptability to new scenarios is an
important step toward robotic decision-making intelligence. To promote the further development of robot autonomous
intelligence, a novel framework RSG is constructed. 
The fundamental skill is a
trained DRL policy network, represented as a node in RSG and linked to specific environment and task nodes. RSG
contains a total of 320 fundamental skills covering a wide distribution of tasks and environments. Essentially, we
define \textit{skills}  ($s$)
 , \textit{environments} ($e$)
 , and \textit{tasks}  $(t)$ as entities and \textit{environment\_to\_skill}  $r_{e \rightarrow s}$
 ,\textit{ task\_to\_skill}  $r_{t \rightarrow s}$ as relations and aim to learn the joint representations of them utilizing KRL and TransH. Once RSG is constructed and represented, it can be flexibly applied to various downstream tasks. When the
matching degree of new skills required by downstream tasks with existing fundamental skills in RSG (i.e., skill
inference scores) is in different ranges, the application mode of RSG also
changes accordingly. In the following sections, we first visualize the construction and representation of RSG. Then we
illustrate RSG's skill inference and execution when the score is high. Finally, we report how to
combine fundamental skills in RSG to rapidly learn new skills when the score is medium or low.

\subsection{RSG Construction and Representation}
Quadruped robot skills, environments, tasks, and their relationships contain rich and valuable information, which is
crucial for robot learning. However, this complex structural information is often hard to model and represent. In this
work, we construct RSG based on two considerations: \textbf{(1) }leveraging \textit{both} prior physics-based knowledge
(e.g., \textit{environment friction, slope, task trajectory}) and latent representations (e.g., \textit{skills} and
\textit{relationships}); \textbf{(2)} encouraging the learning of semantic information: For environment and task nodes,
we encourage RSG to learn from “class” information instead only considering the similarity between individual nodes. Through this approach, we could construct our RSG from both physical prior information and
semantic representation learned from machine learning. 

T-SNE analysis~\cite{47_van2008visualizing} of the representations learned by RSG and baseline is shown in Fig.~\ref{fig_Visual analysis of RSG representation and score assignment.}(\textbf{a-i}). For skill
representations (Fig.~\ref{fig_Visual analysis of RSG representation and score assignment.}\textbf{d}), the raw representation is randomly distributed because we randomly initialize the
skill vectors. After performing RSG training, the learned representations (Fig.~\ref{fig_Visual analysis of RSG representation and score assignment.}\textbf{e}) are separated on both the
task and environment axes. 
\begin{figure}[H]
\centering
\includegraphics[width=0.95\textwidth]{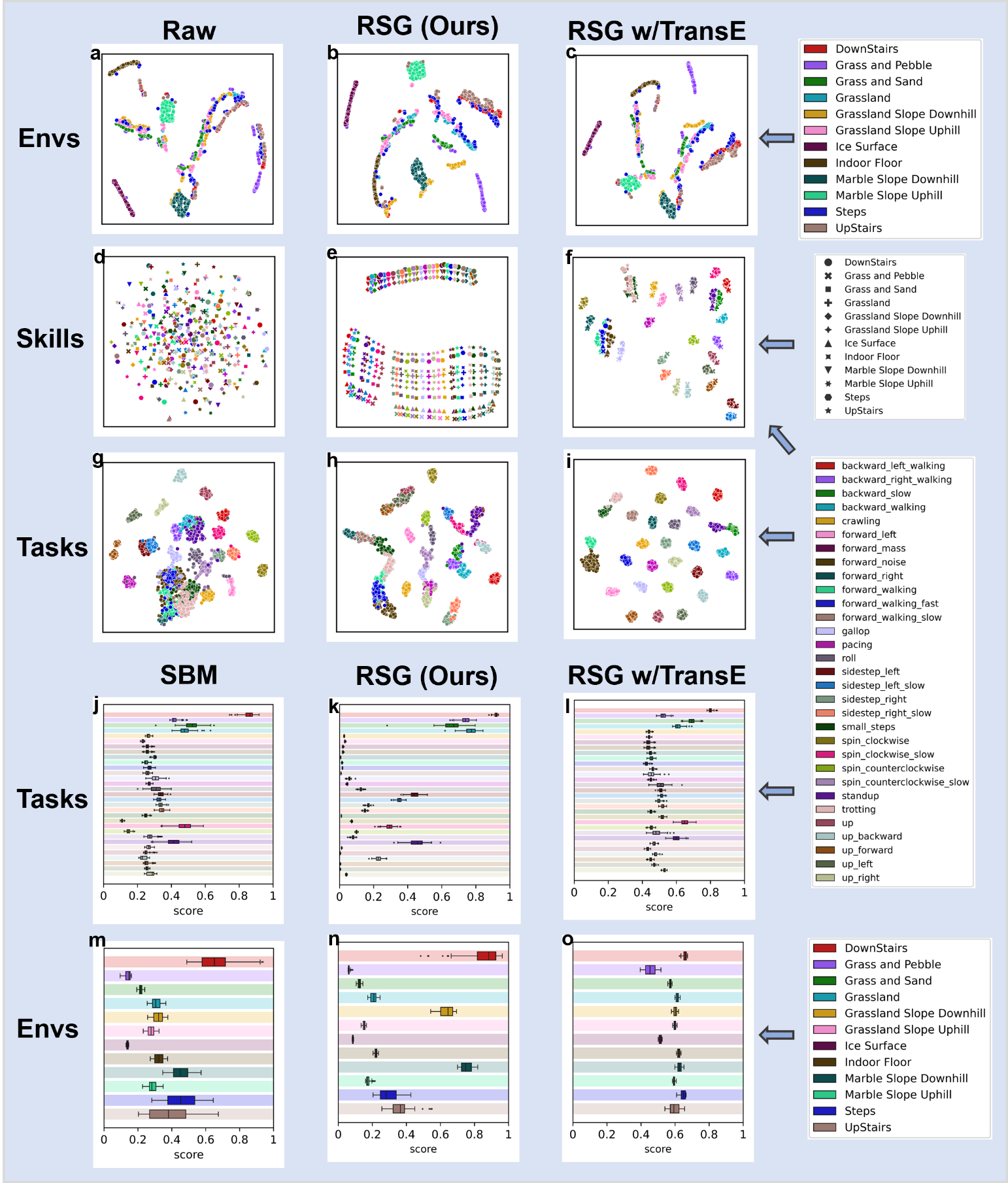}
\caption{\textbf{ Visual analysis of RSG representation and score
assignment.} \textbf{a-i, }T-SNE visualization of environment, skill, and task representations in raw form, RSG (Ours,
w/TransH), and RSG w/TransE, respectively.\textbf{ j-l, }score assignments by SBM, RSG (Ours), and RSG w/TransE for
\textit{backward left walking on grassland }across task queries (31 task categories in total), respectively.
\textbf{m-o,} score assignments by SBM, RSG (Ours), and RSG w/TransE for \textit{forward walking} on
\textit{downstairs} across environment queries (12 environment categories in total), respectively. The box represents
25\% quantile, median, 75\% quantile, and whiskers are data within 1.5 times of the Inter-Quantile Range. The arrows
after each row mark the corresponding legend. Each type is highlighted by the background color of the box plot.
}\label{fig_Visual analysis of RSG representation and score
assignment.}
\end{figure}
In addition, similar environments and tasks are placed in close proximity. 
For example:
\textit{Grassland slope downhill}, \textit{Marble slope downhill,} and \textit{Downstairs}, these similar environments
are put in the upper part while \textit{Ice surface }and \textit{Grass and Pebble }are placed at the bottom. For tasks,
\textit{backward left walking}, \textit{backward right walking}, \textit{backward walking,} and \textit{backward slow}
are placed in the left part\textit{ }while \textit{up backward}, \textit{up left,} and \textit{up forward} are placed
in the right part. For the TransE baseline, due to its inability to model one-to-many, many-to-one, and many-to-many
relationships, the learned representations is only separated on the task axis (Fig.~\ref{fig_Visual analysis of RSG representation and score assignment.}\textbf{f}). 
For task representations, RSG can learn semantic
representations. As shown in Fig.~\ref{fig_Visual analysis of RSG representation and score assignment.}\textbf{g}, the raw representations of “\textit{forward noise}”, “\textit{forward
walking slow}”, “\textit{forward right}”, “\textit{small steps}” and “\textit{forward walking}” are similar and mixed.
In Fig.~\ref{fig_Visual analysis of RSG representation and score assignment.}\textbf{h}, they are successfully separated into clusters with clearer boundaries while still close to each
other, both of which are in line with our expectations. For the results of the baseline TransE (Fig.~\ref{fig_Visual analysis of RSG representation and score assignment.}\textbf{i}), the
tasks are only separated into individual distant clusters without further interaction. For the environment
representations (Fig.~\ref{fig_Visual analysis of RSG representation and score assignment.}\textbf{a-c}), since they can be clearly separated by prior physical information, the raw form
and the learned representations are similar.

Fig.~\ref{fig_Visual analysis of RSG representation and score assignment.}(\textbf{j-o}) shows quantitively how various methods assign scores of a given skill to different
task/environment queries. Fig.~\ref{fig_Visual analysis of RSG representation and score assignment.}(\textbf{j-l}) are obtained by computing the scores of the skill “\textit{backward left
walking on grassland}” for different task queries (including the ground truth “\textit{backward left walking}”). Fig.~\ref{fig_Visual analysis of RSG representation and score assignment.}(\textbf{m-o}) are in a similar setting, except that we compute the scores of “\textit{forward walking on Downstairs}”
for environment queries. The results show that: \textbf{(1)} Purely Similarity-Based Method (SBM, details are in~\ref{app:section_The Similarity-Based Method used in context cross scores}) lacks the ability to consider semantic information and produce more overlapping score
distributions (Fig.~\ref{fig_Visual analysis of RSG representation and score assignment.}\textbf{j, m}); \textbf{(2) }The baseline RSG with TransE cannot distinguish various
environments/tasks, resulting in poor separation (Fig.~\ref{fig_Visual analysis of RSG representation and score assignment.}(\textbf{l, o)});\textbf{ (3)} Our proposed RSG is the only
method that can correctly assign low scores to irrelevant environments/tasks and high scores to relevant
environments/tasks. In addition, the score distributions of different categories have less overlap (Fig.~\ref{fig_Visual analysis of RSG representation and score assignment.}(\textbf{k,
n})). 
More results are in~\ref{app:section_More visual analysis of RSG using t-SNE} and Appendix Fig~\ref{app:fig_Score assignments by RSG for forward walking on various environments.}, \ref{app:fig_Score assignments by RSG for various tasks on Grassland environment.}, \ref{app:fig_Score assignments by SBM for forward walking on various environments.}, \ref{app:fig_Score assignments by SBM for various tasks on Grassland environment.}.

These experiments demonstrate that RSG can effectively model complex graph data of robots and learn
semantically informative representations, thereby condensing robot information into dynamic robot knowledge. RSG can
perform accurate and semantic score assignments, which is critical for reliable skill inference and robust action
execution for downstream tasks.

\begin{figure}[ht]
\centering
\includegraphics[width=0.8\textwidth]{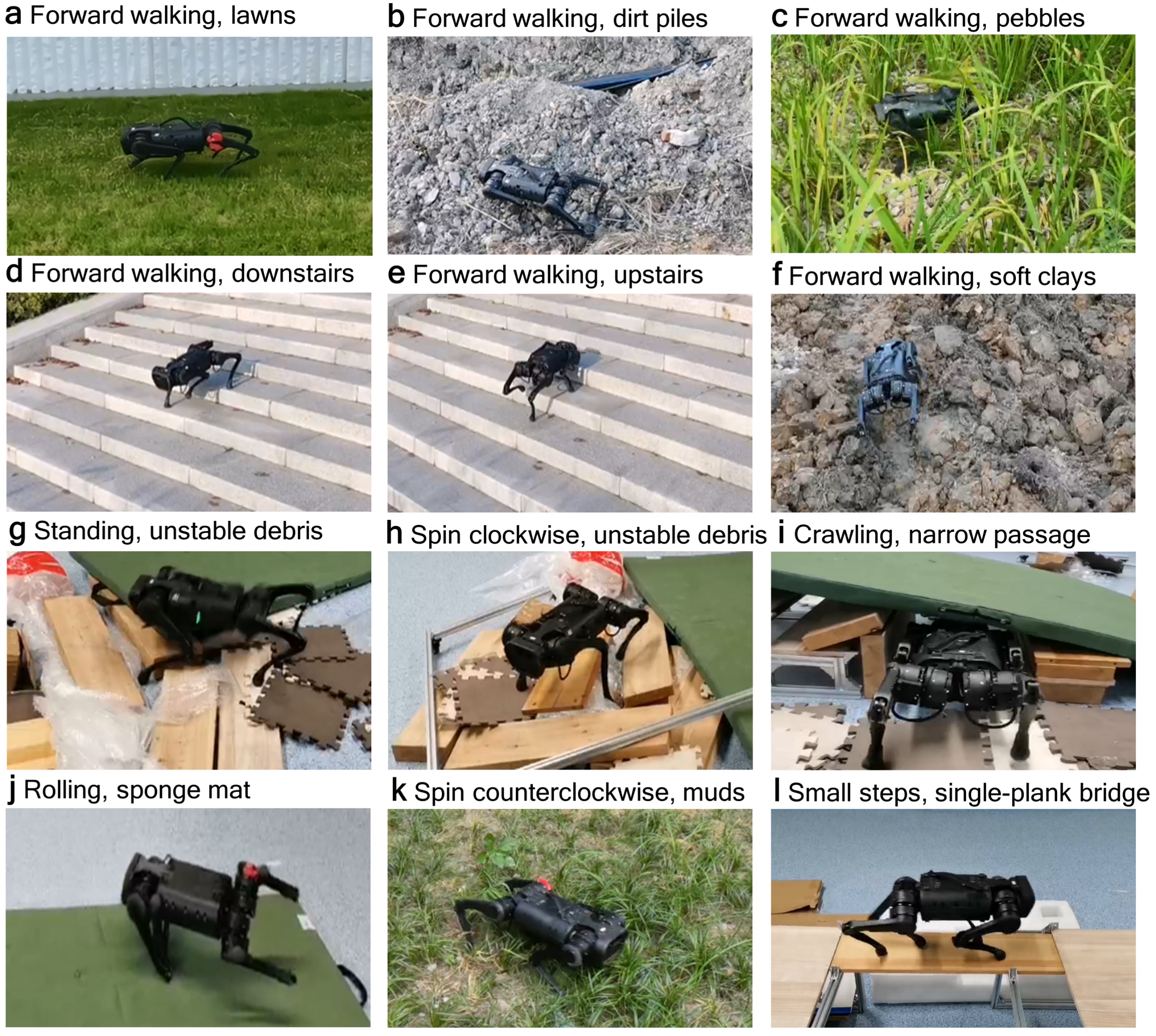}
\caption{\textbf{Broad distribution of fundamental skills in RSG.}
Robot tasks include rolling, standing, walking, turning, small steps, etc. The environment includes unstable debris,
narrow passages, single-plank bridges, sponge mats, dirt piles, lawns, mud, pebbles, stairs, soft clays, etc.
}
\label{fig_Broad distribution of fundamental skills in RSG.}
\end{figure}
\subsection{Skill Query, Inference and Reuse}
\textbf{Diversity fundamental skills in RSG.}
RSG has a wide range of skills, with a total of 320 fundamental skills, covering 12 environments with
different terrain features and 31 flexible and diverse tasks. Some unreasonable skills are
excluded, such as rolling on stairs. The real machine deployment of some fundamental skills is shown in Fig.~\ref{fig_Broad distribution of fundamental skills in RSG.} and
Video 1. 
The quadruped robot equipped with RSG can perform various behaviors in reality. 
As shown in Fig.~\ref{fig_Broad distribution of fundamental skills in RSG.}(\textbf{a-f}), the quadruped robot can move robustly and successfully traverse
various terrains. 
In Fig.~\ref{fig_Broad distribution of fundamental skills in RSG.}(\textbf{g-l}), the robot demonstrates a variety of
complex behavioral skills. These diverse and highly dynamic behaviors are effectively organized under the RSG
framework, laying a solid foundation for skill inference and rapid skill learning.

\begin{figure}[H]
\centering
\includegraphics[width=0.9\textwidth]{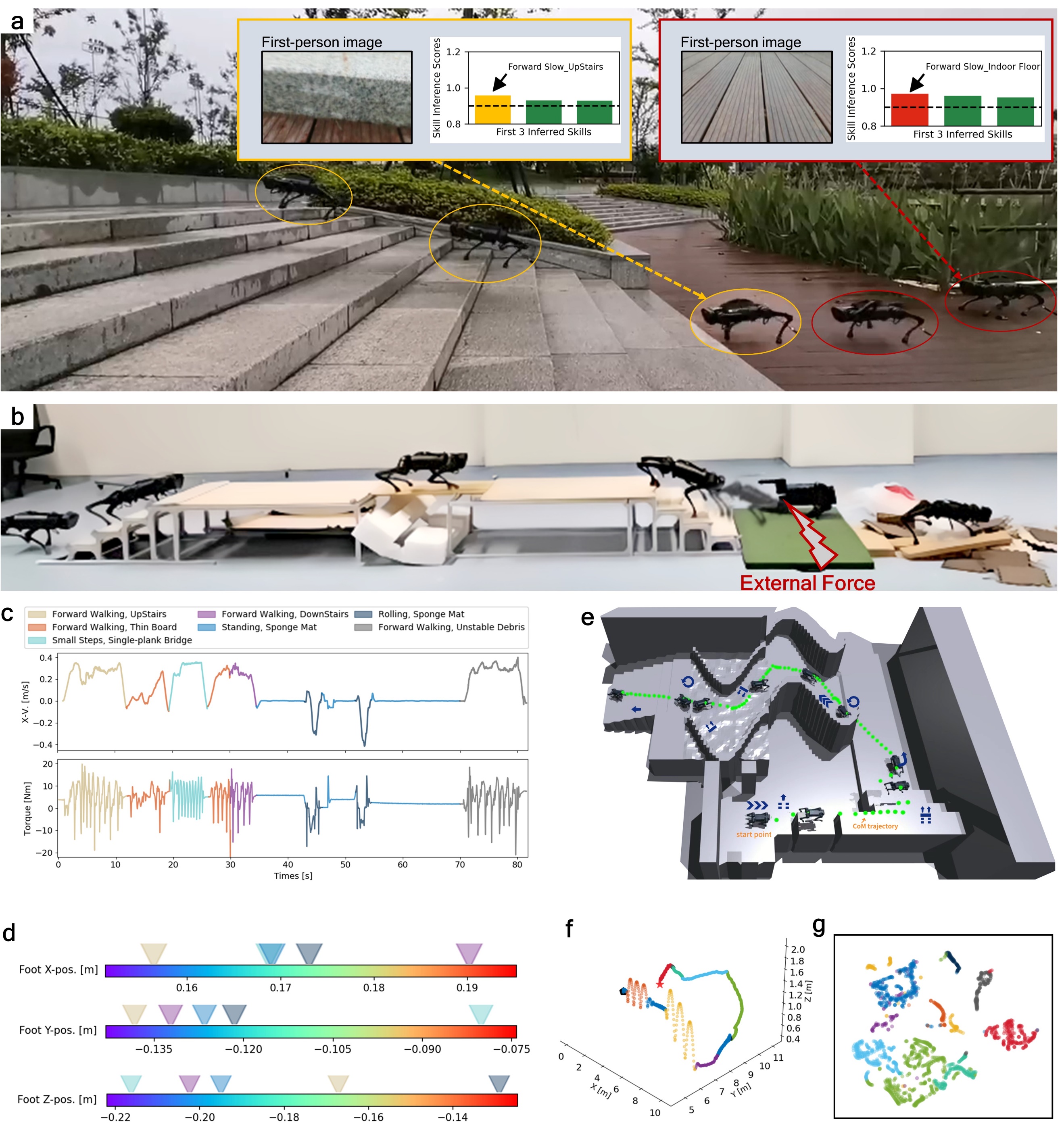}
\caption{\textbf{The deployment of RSG for robot parkour.} \textbf{a},
Autonomous inference of real-world robot skills. Two inferred skills and the corresponding first-person images are
shown. \textbf{b}, Robust execution of diverse robot skills in a realistic parkour task. \textbf{c}, The profiles of
robot body velocity and right front leg torque corresponding to \textbf{b}. \textbf{d}, Average of foot position (right
front leg) trajectories corresponding to several skills in \textbf{b}. \textbf{e-g}, Ten snapshots of robot motion
in simulation, CoM position trajectories, and t-SNE visualizations of the executed skills.
}
\label{fig_The deployment of RSG for robot parkour.}
\end{figure}

\textbf{Skill inference and execution.}
To cope with complex and unstructured scenarios, robot autonomous intelligence requires a large number of flexible
behaviors and efficient skill inference. In this section, to test the RSG's autonomous skill
selection and discovery, skill reliable inference, and robust action execution capabilities when the skill inference
score is high (greater than 0.9), three groups of experiments are designed: outdoor open scene (Fig.~\ref{fig_The deployment of RSG for robot parkour.}\textbf{a}),
indoor parkour scene (Fig.~\ref{fig_The deployment of RSG for robot parkour.}(\textbf{b-d)}) and simulated parkour task (Fig.~\ref{fig_The deployment of RSG for robot parkour.}(\textbf{e-g)}).

In the outdoor open scene (Fig.~\ref{fig_The deployment of RSG for robot parkour.}\textbf{a}), we utilized the multimodal Large Language Model (LLM) Gemini-1.5-pro~\cite{48_team2024gemini}
to achieve real-time perception and understanding of environmental information and real-time acquisition of
environmental queries. By designing appropriate prompts, Gemini-1.5-pro can utilize visual information to convert the
terrain ahead into environmental queries for the skills required. After receiving the queries, RSG can autonomously
infer the best matching skill and execute it. Specifically, after the user gives the behavior of interest (forward
movement), the robot can infer the matching skills (\textit{Forward Slow\_Indoor Floor} and \textit{Forward
Slow\_UpStairs}) based on real-time environmental perception to complete the specified task. More experimental details
are in~\ref{app:section_The autonomous query of new environments} and Video 2.

In the indoor parkour scene (Fig.~\ref{fig_The deployment of RSG for robot parkour.}(\textbf{b-d}) and Video 3), we complete the extraction and execution of skills
through pre-designed environment and task queries. We pre-plan the robot's expected trajectory and
infer the corresponding skills from RSG in parkour to complete the given task trajectory. The characteristics of the
various skills performed during parkour are precisely described in Fig.~\ref{fig_The deployment of RSG for robot parkour.}(\textbf{c-d}). In particular, this task is
challenging because it not only contains various unstructured scenarios, but also involves the robot falling due to
external disturbances. In this case, RSG is able to accurately extract appropriate skills (rolling, posture recovery,
walking, etc.) and robustly execute them to continue to complete the given task. It is difficult to integrate these
different skills into the parkour task with a single general policy, and therefore it is difficult to cope with
external disturbances. 

For the simulated parkour task (Fig.~\ref{fig_The deployment of RSG for robot parkour.}\textbf{e-g}), we designed several more challenging terrains and tasks. The
presence of continuous forward and leftward jumping actions poses additional challenges to body direction control and
foot placement. The robot must be able to accurately infer fundamental skills and perform them flexibly and stably to
traverse a variety of terrains with large CoM variations (Fig.~\ref{fig_The deployment of RSG for robot parkour.}\textbf{f}). The t-SNE visualization results in Fig.~\ref{fig_The deployment of RSG for robot parkour.}\textbf{g} show that the fundamental skills performed in RSG are clearly differentiated. More results are in
Appendix Fig.~\ref{app:fig_The profiles of two front feet positions.} and Video 4.

These experiments show that RSG has the ability of autonomous skill selection and discovery, reliable skill inference,
and robust action execution to cope with complex and diverse unstructured terrains.


\subsection{Rapid Skill Learning}
Animals can quickly learn new skills by trying existing behaviors several times. In this section, we set up three sets
of experiments to analyze RSG's ability to rapidly adapt to new scenes and quickly learn new
skills when the skill inference score is medium (between 0.7 and 0.9) and low (less than 0.7): jumping to form a circle
(JFC) (Fig.~\ref{fig_Rapid adaptation to new scenarios, and fast learning of unseen new skills.}(\textbf{a-d})), balancing on a slope (Fig.~\ref{fig_Rapid adaptation to new scenarios, and fast learning of unseen new skills.}(\textbf{e-g})) and learning of complex new skills
corresponding to random hand-drawn trajectory sketches (Fig.~\ref{fig_Rapid adaptation to new scenarios, and fast learning of unseen new skills.}(\textbf{h-j})).

In the new downstream task of JFC (Fig.~\ref{fig_Rapid adaptation to new scenarios, and fast learning of unseen new skills.}\textbf{b}), the skill inference of RSG is shown in Fig.~\ref{fig_Rapid adaptation to new scenarios, and fast learning of unseen new skills.}\textbf{a}. The top
three most matched fundamental skills inferred are \textit{jump in place (JP)\_indoor floor}, \textit{jump left
(JL)\_indoor floor}, and \textit{jump backward (JB)\_indoor floor}. The unspecified fundamental skill environment is
the \textit{indoor floor}, the same applies hereafter. In the RSG framework, the top three fundamental skills with
medium skill inference scores are utilized to rapidly adapt to new scenes through Bayesian Optimization (BO) on the
skill action space. The trend of the change of bias and weight during the BO composition
process of this downstream task is shown in Fig.~\ref{fig_Rapid adaptation to new scenarios, and fast learning of unseen new skills.}\textbf{c}, and the results show that it can converge within dozens
of iterations. Intuitively, the basic properties of the new skill JFC can be implicitly encoded by the fundamental
skills JP, JL, and JB, and achieved by combining them together. The results in Fig.~\ref{fig_Rapid adaptation to new scenarios, and fast learning of unseen new skills.}\textbf{d} also confirm this
assertion, where the composition utilizing BO achieves a performance level similar to that of training from scratch
with the PPO algorithm~\cite{49_schulman2017proximal}. Furthermore, the initial weight of BO composition is determined by the skill inference
score obtained from RSG. This approach has been confirmed by the ablation experiment in Fig.~\ref{fig_Rapid adaptation to new scenarios, and fast learning of unseen new skills.}\textbf{d} to promote the
robot's “warm boot”. More results in Video 5.

To verify RSG's ability to rapidly adapt to new environments, we compared several methods for
maintaining balance in a slope environment outside the training distribution. Qualitative (Fig.~\ref{fig_Rapid adaptation to new scenarios, and fast learning of unseen new skills.}\textbf{e}) and
quantitative (Fig.~\ref{fig_Rapid adaptation to new scenarios, and fast learning of unseen new skills.}\textbf{f}) comparison results show that the BO composition with inferred skills can significantly
enable the robot to have the smallest body posture fluctuation range, thereby successfully maintaining balance on a
steep slope. Moreover, as the slope increases, the weight of the fundamental skill standing becomes smaller (Fig.~\ref{fig_Rapid adaptation to new scenarios, and fast learning of unseen new skills.}\textbf{g}). This shows that the BO with inferred skills in RSG can adjust the action weights of fundamental skills
according to the characteristics of the environment, thereby better adapting to the new environment. More new
environment adaptation experiments are in Video 6, and the rapid learning processes of the new gait are in
Appendix Fig.~\ref{app:fig_The novel realistic quadrupedal gait rapid learning processes.} and Video 7. 
\begin{figure}[H]
\centering
\includegraphics[width=0.9\textwidth]{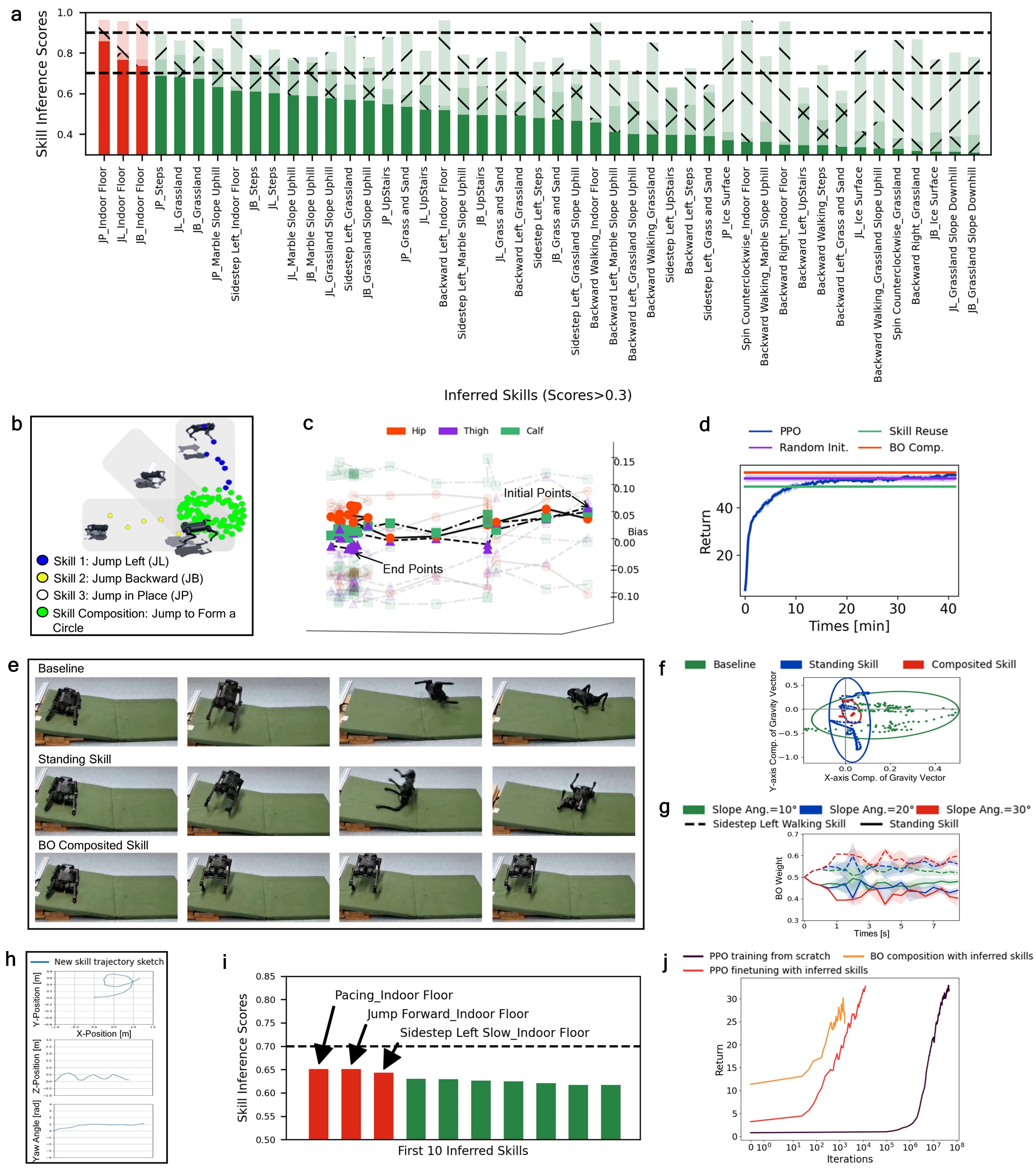}
\caption{\textbf{Rapid adaptation to new scenarios, and fast learning of unseen new skills. a}, Distribution of
skill inference scores in RSG corresponding to \textbf{b}. The values \hspace{0pt}\hspace{0pt}corresponding to the
highest points of labels “/” and “{\textbackslash}” indicate the degree of match with the task and environment queries,
respectively. \textbf{b}, Overview of the process of rapid adaptation to the new task. \textbf{c}, Trends of BO weights
and biases corresponding to the fundamental skill (right hind leg) of \textbf{b}. The shaded area indicates the trends
of other legs. \textbf{d}, Performance comparison with baseline.\textbf{ e-f}, Comparison of rapid adaptation to new
realistic environments, and the corresponding comparison of the projection of the gravity vector on the horizontal
plane (the ellipse represents 2.5 standard deviations). \textbf{f}, Trends of BO weights under adaptation to different
new terrain slopes.\textbf{ h}, Manually drawn robot trajectory sketch, which is smoothed as a task query for a
required unseen new skill. \textbf{i-j}, Inferred skills in RSG and fast learning curves corresponding to
\textbf{h}.
}
\label{fig_Rapid adaptation to new scenarios, and fast learning of unseen new skills.}
\end{figure}

Furthermore, we consider the task of fast skill learning of the robot under the condition of low skill inference score
(Video 8). Fig.~\ref{fig_Rapid adaptation to new scenarios, and fast learning of unseen new skills.}\textbf{h} shows a complex manually drawn robot trajectory sketch, which is smoothed and utilized as
a task query for downstream tasks. The highest inference score of RSG is less than 0.7 (Fig.~\ref{fig_Rapid adaptation to new scenarios, and fast learning of unseen new skills.}\textbf{i}), so the first three inferred skills are fine-tuned online with PPO.
The results (Fig.~\ref{fig_Rapid adaptation to new scenarios, and fast learning of unseen new skills.}\textbf{j}) show that the performance of PPO fine-tuning using the
inferred skills is comparable to that of PPO training from scratch, but the number of fine-tuning steps is only
slightly more than 10,000 steps (about 5 minutes), which greatly improves the sample efficiency of new skill learning.
Moreover, the BO composition requires about 1,000 iterations (about 1 minute), but the final return fluctuates between 25 and 30. Therefore, for downstream tasks with low
skill inference scores, utilizing PPO fine-tuning based on the inferred fundamental skills is the best choice. More
skill adaptation experiments corresponding to hand-drawn trajectory sketches with medium skill inference scores
in Video 9.

The experiment shows that the robot configured with RSG has the ability to rapidly adapt to new scenarios and learn
skills quickly, which is an indispensable part of robot intelligence and autonomy.

\section{Conclusion}
Lack of behavioral inference and fast dynamic adaptation in the face of complex and diverse scenarios is one of the key
challenges that hinder the rapid development of quadruped robot motion control in the physical world. Our research is
dedicated to exploring the flexible and adaptive motion behaviors of quadruped robots, which can generate versatile
motions from existing basic skills according to changes in tasks and environments. We propose a novel framework RSG
that links tasks and environments with motion skills and allows many different abilities to be associated based on the
connections between them. RSG can be utilized to organically organize, query, infer, and execute a large number of
skills of quadruped robots. Its key advantages are mainly reflected in skill diversity, robust execution of skills,
autonomous skill selection, rapid adaptation, and fast learning. Our work has taken an important step in the autonomous
inference and decision-making of robots for unstructured scenes and complex tasks. Therefore, the significance of our
work is that it provides a feasible framework for the flexibility and adaptability of robot behavior and lays a solid
foundation for promoting the autonomous intelligence of robots.


\section*{Declaration of competing interest}
The authors declare that they have no known competing financial interests or personal relationships that could have appeared to 
influence the work reported in this paper.

\section*{Data availability}
The realistic demonstration data for AMP training are available at
\url{https://drive.google.com/drive/folders/1iriIEETEtjkZZnaF9BHNSoXFocyUZfVv?usp=sharing}.

\section*{Code availability}
The code is available at \url{https://github.com/COST-97/RSG}.

\section*{Acknowledgments}
This work was supported by the National Science and Technology Innovation 2030 - Major Project (Grant No. 2022ZD0208800), and NSFC General Program (Grant No. 62176215).

\appendix
\section{Data collection details, observations, and reward functions}
\label{app:section_Data collection details, observations, and reward functions}
The hardware platform used in this work is the \textit{Unitree A1} quadrupedal robot, which has a built-in controller
that can follow high-level motion commands. We used a joystick to send velocity commands and generate multiple expert
trajectories in various realistic terrains. The expert trajectories consist of base linear velocity, base angular
velocity, joint position, joint velocity, body height, and foot position relative to the body. These demonstrations
provide diverse periodic motion behaviors that can guide control policy training. The data were collected every 0.02 s
using the legged SDK from the \textit{Unitree A1} robot. At each time step, the information of contact force sensors,
IMU, and joint encoders are recorded. 

We utilize an asymmetric actor-critic architecture for policy training. Actions are defined as desired joint positions
of 12 actuators at 50 Hz. The desired joint angles were tracked using the PD controller at 200 Hz ( $K_p=28$
 and  $K_d=0.7$
 ). The observation space of the policy network includes projected gravity, joint positions, joint velocities, actions
of the previous time step, body angular velocity, and body linear velocity on the XY axis. The input of the critic
network contains the policy observation, height scan around the body, foot contact forces, foot contact states,
foot-ground friction coefficients, payloads, and proportional and derivative gains of the PD controller.

Inspired by previous works~\cite{25_nahrendra2023dreamwaq,59_rudin2022learning,66_margolis2023walk}, the task reward is a linear combination of basic reward functions multiplied by
the time step  $\triangle t=0.02s$
 . The notations used in 
Appendix Tab~\ref{app:Name and definition of reward functions.} are summarized as follows: $\exp (\cdot)$ and $\operatorname{var}(\cdot)$ are exponential and variance operators, respectively. $(\cdot)^{\text {des }}$ and $(\cdot)^{\text {cmd }}$ indicate the desired and commanded values, respectively. $x, y$, and $z$ are defined on the robot's body frame, with $x$ and $z$ pointing forward and upward, respectively. $g, v_{x y}, \omega_{\mathrm{yaw}}, \theta_{\text {base }}, p_{f, y, j}, p_{f, z, j}, v_{f, x y, j}$ and $\tau$ are the gravity vector projected into the robot's body frame, linear velocities, yaw angle velocity, body orientation, foot width, foot height, foot lateral velocity, and joint torque, respectively. $a_t, a_{t-1}$ and $a_{t-2}$ are the actions of the current time step $t$, the previous time step $t-1$ and $t-2$ respectively. $h_{\text {base }}$ and $h_{\text {target }}$ respectively represent the current body height and expected body height of the robot. $g_{\text {default }}=[0,0,-1]$ is the default gravity vector. $\mathbf{1}_{\text {contact}, j}$ denotes the contact state of the $j^{\text {th}}$ leg. $t_{air,j}$ denotes the swing time of the $j^{\text {th}}$ leg in the air. The desired contact states $C_j^{cmd}(\cdot)$ are defined in previous work~\cite{66_margolis2023walk}.

\section{Skill composition by using Bayesian optimization}
\label{app:section_Skill composition by using Bayesian optimization}
In this section, we describe how fundamental skills are composed by using the BO method. Without loss of generality,
suppose two different fundamental skills are selected from the RSG. The actions of the corresponding control policy are
Gaussian distribution, which can be expressed as:
\begin{equation}
A_1 \sim \mathcal{N}\left(\mu_1, \sigma_1\right), A_2 \sim \mathcal{N}\left(\mu_2, \sigma_2\right),
\end{equation}
where $\mu_i$ and $\sigma_i$ are the mean and standard deviation of the Gaussian distribution, respectively. The goal of skill composition is to find a linear combination of two skills:
\begin{equation}
A_{\text {new}}=k_1 A_1+k_2 A_2+b \sim \mathcal{N}\left(k_1 \mu_1+k_2 \mu_2+b,\left(k_1 \sigma_1\right)^2+\left(k_2 \sigma_2\right)^2\right),
\end{equation}
that can maximize the cumulative rewards on the new task or environment.

We utilized BO as the search method for the parameters $\mathbf{x}=\left[k_1, k_2, b\right]$. Assume that the observed return $y=f(\mathbf{x})+\epsilon$ is an objective function of the parameters. $f(\mathbf{x})$ is described by a Gaussian Process (GP) model:
\begin{equation}
f(\mathbf{x}) \sim \mathcal{G P}\left(m(\mathbf{x}), k\left(\mathbf{x}, \mathbf{x}^{\prime}\right)\right),
\end{equation}
where $m(\mathbf{x})$ and $k\left(\mathbf{x}, \mathbf{x}^{\prime}\right)$ are the mean function and kernel function, respectively. $\epsilon$ is the measurement noise of the returns, which is assumed to follow a normal distribution $\epsilon \sim \mathcal{N}\left(0, \sigma_{\text {noise }}^2\right)$, where $\sigma_{\text {noise }}^2$ is a hyperparameter. the mean function is a constant $m_0$ and the kernel function is selected as a Gaussian kernel:
\begin{equation}
k\left(\mathbf{x}, \mathbf{x}^{\prime}\right)=\sigma_f^2 \exp \left(-\frac{1}{2 l^2}\left\|\mathbf{x}-\mathbf{x}^{\prime}\right\|^2\right),
\end{equation}
where $\sigma_f$ and $l$ are the hyperparameters of the kernel. The sampled parameters are stored in a dataset $D$ and the GP model is iteratively updated to fit the dataset and find the optimal point $\mathbf{x}^*$.

\section{The optimization target for BO composition and PPO finetuning}
\label{app:section_The optimization target for BO composition and PPO finetuning}
The BO and PPO methods used for medium and low inference scores respectively use the same reward function form. The
optimization target is a linear combination of different reward function terms: 
\begin{equation}
\begin{aligned}
& R_{target}=5 L V T\left(v_{x y}, v_{x y}^{cmd}\right)+1.5 A V T\left(\omega_{yaw}, \omega_{yaw}^{cmd}\right)+0.3 L O \\
& +0.3 L O R P Y-0.3 F F C+0.3 J B H-0.003 A R-0.00003 T S,
\end{aligned}    
\end{equation}
including robot CoM linear velocity and angular velocity tracking (LVT, AVT), posture stability (LO, LOPRY), foot contact (FFC), jump (JBH) and action smoothness (AR, TS). The specific mathematical definitions are in Appendix Tab~\ref{app:Name and definition of reward functions.}.

The 11 CoM trajectories of the new skills to be learned are used as queries to perform skill inference and extract the most suitable fundamental skills in RSG. Meanwhile, they are also used as a periodic behavior of the new skill, that is, with a period of 11 in time steps, to determine the values of $v_{x y}^{c m d}$ and $\omega_{y a w}^{c m d}$ in the optimization target $R_{target}$.




\section{The alleviation of the sim2real problem}
\label{app:section_The alleviation of the sim2real problem}
Alleviating the simulation-to-reality gap is pivotal for the effective deployment of learning-based robotic
systems in real-world scenarios. Presently, there exists no universally applicable solution to completely bridge this
gap. During the DRL training of fundamental skills, many design details may potentially affect the sim2real gap. Here
we mainly consider five aspects: \textbf{(1)} expert data collection,  \textbf{(2)} algorithm framework,  \textbf{(3)} environment hyperparameter
design,  \textbf{(4)} reward and loss design, and  \textbf{5)} skill inference robustness. Accordingly, in this work, we systematically
applied the following methods to alleviate sim2real gaps:
\begin{enumerate}[series=listWWNumxix,label=\textbf{(\arabic*)},ref=\arabic*]
\item {
In terms of expert data collection, we used a joystick to control the movement of the real robot so that the AMP
algorithm~\cite{50_escontrela2022adversarial} could be used to learn the fundamental skills of the expert style. When using the joystick to control,
it is necessary to ensure that the robot's movement is as smooth as possible, so that its CoM speed is maintained at a
fixed value, which is used as the desired value of the task-related reward during AMP training.}
\item {
In terms of DRL training of fundamental skills, two key algorithm designs make the skills more robust: (\textbf{2.1)} We
use an asymmetric Actor-Critic framework, that is, the actor policy can only receive proprioception, but the Critic
network can also receive privileged information that is difficult to obtain in reality. This design ensures better
performance of the policy and can stably deploy the trained actor policy to a real machine without the need for a
teacher-student framework~\cite{24_lee2020learning}. that would degrade performance; (\textbf{2.2) }a context-assisted estimation network
(CEN) is used to improve the accuracy of CoM linear velocity estimation and the robustness of decision-making actions
by jointly estimating body state and environmental context. There have been some previous studies on learning latent
representations of terrain characteristics to enhance the robustness of robot actions~\cite{67_kumar2021rma,68_fu2021minimizing,69_margolis2024rapid}. In addition, using a
learning network to estimate CoM linear velocity can also significantly improve the robustness of the control policy~\cite{25_nahrendra2023dreamwaq,70_ji2022concurrent}
because it can eliminate the accumulated estimation drift.}
\item {
A common domain randomization method in DRL-based quadruped motion control tasks is adopted, including uniform sampling
of physical parameters, random application of disturbance forces to the body, and addition of noise to observations~\cite{25_nahrendra2023dreamwaq,50_escontrela2022adversarial}. Specifically, during the fundamental skill training process, the base mass difference and PD gains are
uniformly sampled in the intervals of $[-1,1]$ and $[0.9,1.1]$ respectively. Meanwhile, the robot is subjected to a random
thrust in the horizontal direction every $7s$. Emulates an impulse by setting a randomized base velocity (speed range
$[0,1]$). The body observations obtained in the simulation will add a certain degree of noise to simulate real sensors,
including joint positions, joint angular velocities, linear velocities, angular velocities, gravity and height
measurements (0.03, 1.5, 0.1, 0.3, 0.05 and 0.1).}
\item {
For the design of the reward and loss function of DRL (and AMP) for training fundamental skills, some auxiliary terms
are also helpful~\cite{26_yang2020multi}. 
There are also two auxiliary items in the task-related reward function to help alleviate the
sim2real problem. They punish the huge torque caused by large actions and the violent fluctuations between actions,
thereby ensuring low energy consumption and smoothness of the robot's action. For some fundamental skills that are
highly dynamic (rolling, standing up after rolling, etc.), to encourage more gentle and effective behaviors, we add
regularization terms to the loss function of PPO to make the action close to the given joint position. The given joint
position is the key joint position in the robot's motion process that is set manually. Compared with setting related
terms in the reward function, adding regularization terms to the loss function is a more direct and effective way.}
\item {
To boost the inference robustness of RSG for new environment and task queries, we carefully designed the input data to
retain the “class” information. That is, RSG could utilize information in both property similarity and “class”
information when queried by a specific environment or task. }
\end{enumerate}

\section{The Similarity-Based Method used in context cross scores}
\label{app:section_The Similarity-Based Method used in context cross scores}
To compute context cross scores by SBM methods, we first calculate the centroid of original context class, then compute
the similarity of each new context instance to it:
\begin{equation}
    \kappa(c_n,c_o) = \text{norm}_c(\vert\vert c_n - \text{class}(c_o) \vert\vert),
\end{equation}
where  $\text{class}(\cdot)$ finds the centroid of class. $\text{norm}_c(x) = x/x_{\text{max}}$ and   $x_{\text{max}}$ are computed over all samples in context $c$. Then the similarity score is calculated: $s(c_n, c_o)=\exp(\tau \times \kappa(c_n, c_o))$, where $\tau=3$ is a temperature hyperparameter.

\section{More visual analysis of RSG using t-SNE}
\label{app:section_More visual analysis of RSG using t-SNE}
\begin{figure}[H]
\centering
\includegraphics[width=0.95\textwidth]{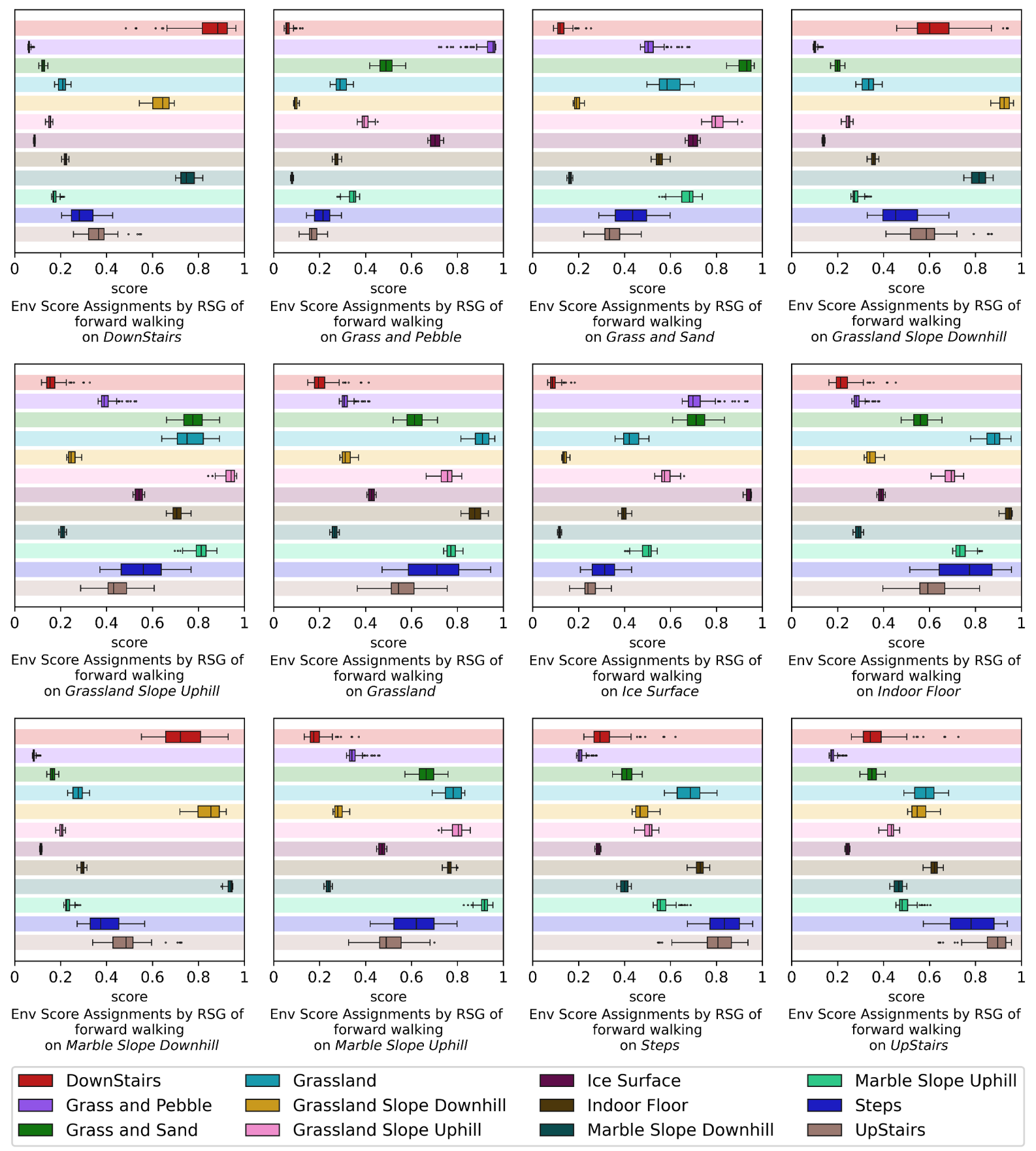}
\caption{
Score assignments by RSG for forward walking on various environments.
}
\label{app:fig_Score assignments by RSG for forward walking on various environments.}
\end{figure}

\begin{figure}[H]
\centering
\includegraphics[width=0.95\textwidth]{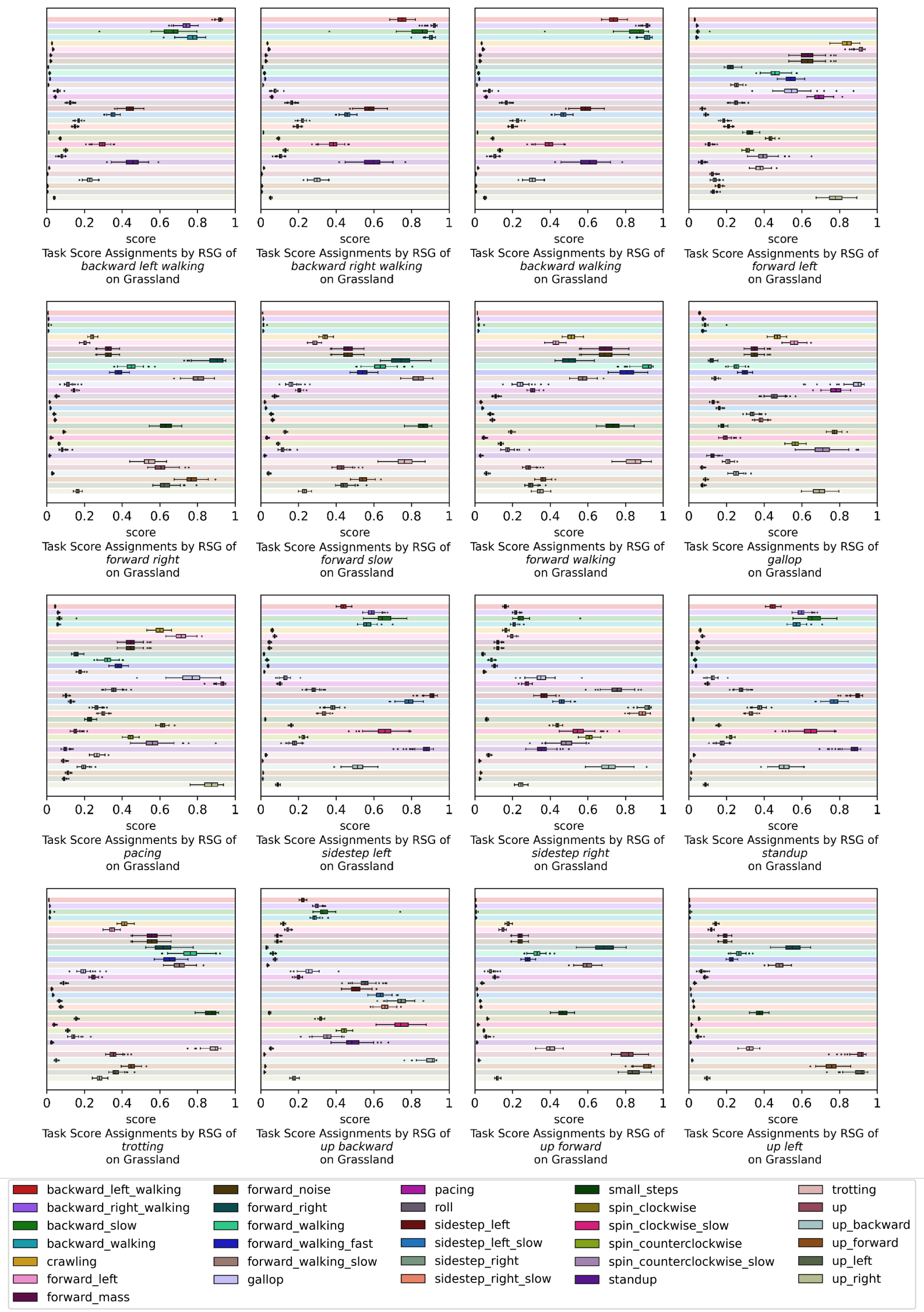}
\caption{
Score assignments by RSG for various tasks on Grassland
environment.
}
\label{app:fig_Score assignments by RSG for various tasks on Grassland environment.}
\end{figure}

\begin{figure}[H]
\centering
\includegraphics[width=0.95\textwidth]{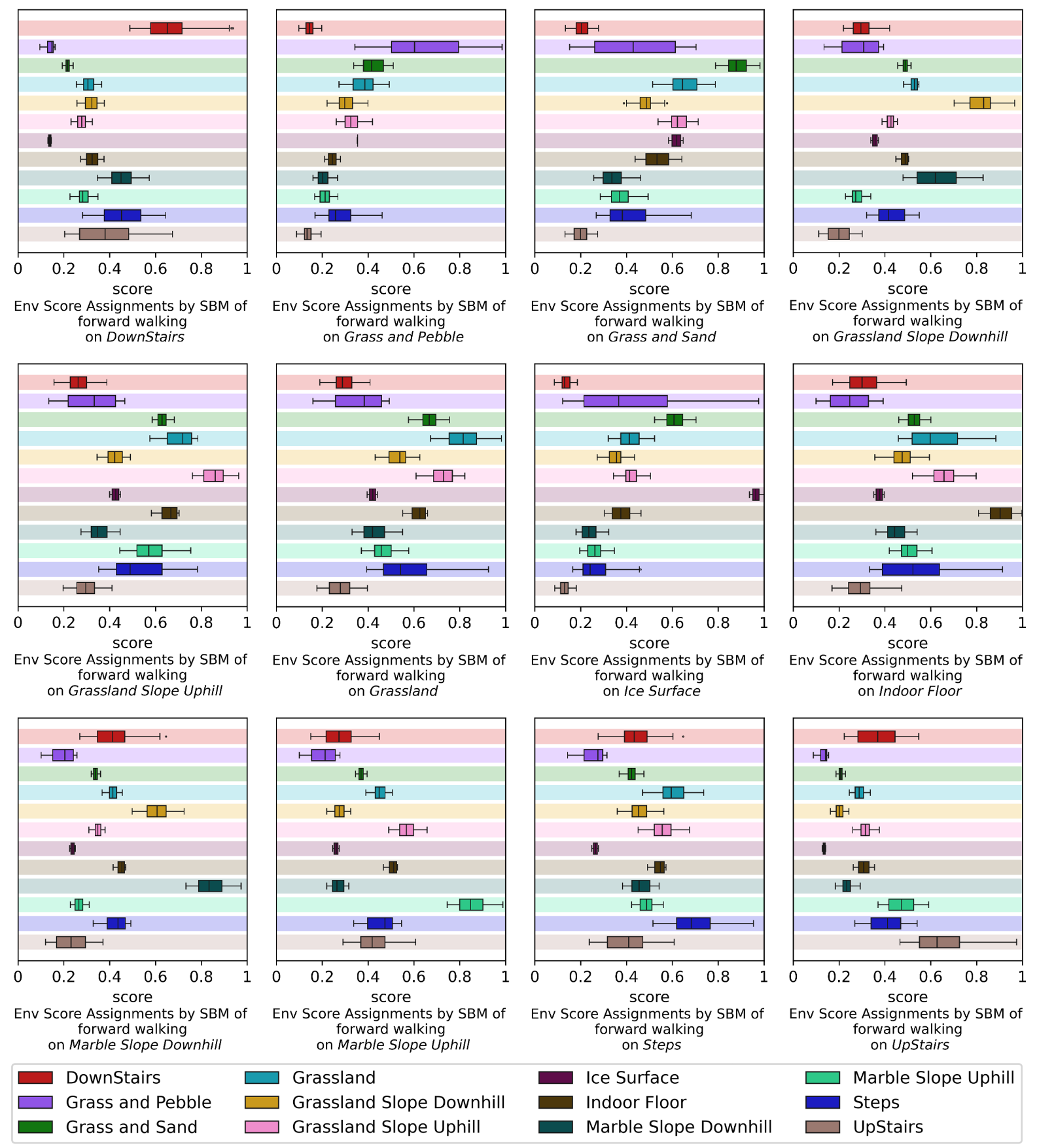}
\caption{
Score assignments by SBM for forward walking on various environments.
}
\label{app:fig_Score assignments by SBM for forward walking on various environments.}
\end{figure}

\begin{figure}[H]
\centering
\includegraphics[width=0.95\textwidth]{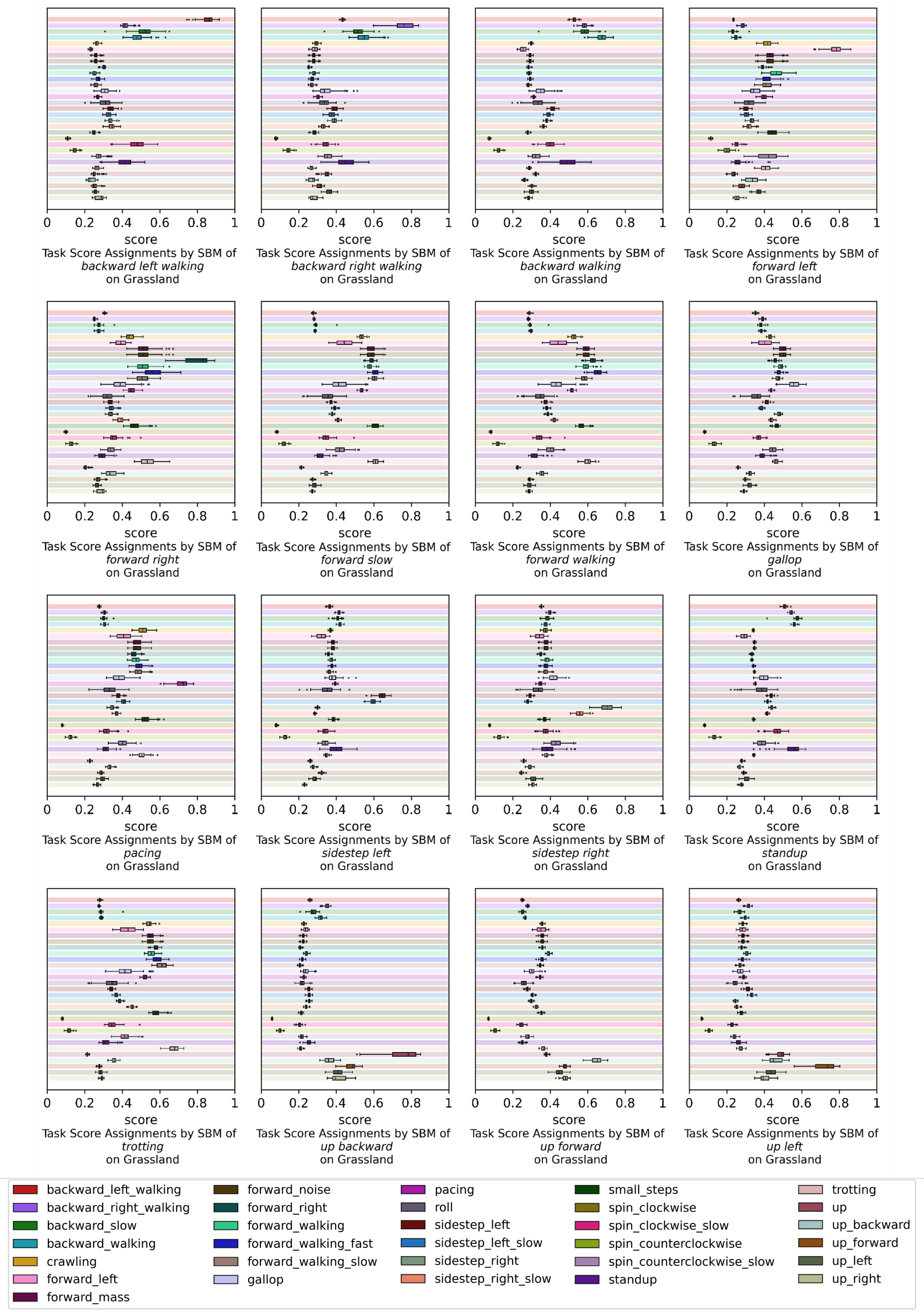}
\caption{
Score assignments by SBM for various tasks on Grassland environment. 
}
\label{app:fig_Score assignments by SBM for various tasks on Grassland environment.}
\end{figure}

Fig.~\ref{app:fig_Score assignments by RSG for forward walking on various environments.} shows that environment cross scores of \textit{forward walking} in all possible environments. 
Fig.~\ref{app:fig_Score assignments by RSG for various tasks on Grassland environment.} shows task
cross scores on \textit{grassland} for all possible tasks. Similar results are found on other tasks or environments,
which are omitted for the sake of brevity. The visualization results show that the RSG outputs scores with the
following two properties: \textbf{(1) }Non-related contexts are usually scored close to zero while highly-related
contexts are scored near one. \textbf{(2)} The score ranges of different context clusters are separated by clearer
boundaries, which means that RSG discovers semantic representations. 
Furthermore, the results in Fig.~\ref{app:fig_Score assignments by SBM for forward walking on various environments.} and \ref{app:fig_Score assignments by SBM for various tasks on Grassland environment.} show
that the context cross scores by using the SBM method usually lack semantic information and span more widely. The
scores for self-context (the original context used to train the skills) are not always close to one, and the scores for
obviously irrelevant contexts are not close to zero. Therefore, vanilla similarity-based methods do not utilize
predefined labels and struggle to learn clearly separated representations.

\section{The autonomous query of new environments }
\label{app:section_The autonomous query of new environments}
For the autonomous query of new environments, we can use trajectory sketches to support the convenient
acquisition of new task queries, and use the multimodal large language model Gemini-1.5-pro \cite{48_team2024gemini} to achieve perception
and understanding of environmental information, as well as real-time acquisition of environmental queries. By designing
appropriate prompts, the Gemini-1.5-pro can use visual information and classify the terrain in front as one of the
environments given in this work, and output the corresponding attribute range. After receiving the query for a new
skill, RSG can autonomously infer the best matching skill and execute it. The system and user are two types of user
queries and Gemini represents the reply of Gemini. Specific prompts used when querying Gemini-1.5-pro:

\begingroup
\scriptsize
\textbf{\textit{System}}\textit{: I will send some terrain images to you. You need to classify these terrains into
corresponding categories and infer their properties. The terrain categories and their corresponding properties
(friction coefficients (the larger, the larger friction for walking on this terrain), flatness (the larger, more
un-uniform this terrain is) and slope (the larger, the more steep this terrain is)) are as followings:}

\textit{Indoor Floor: friction: 0.6 to 0.9, flatness: 0.0, slope: 0.0}

\textit{Ice Surface: friction: 0.01 to 0.1, flatness: 0.0, slope: 0.0}

\textit{UpStairs: friction: 1.2 to 1.5, flatness: 0.0 to 13.125, slope: 0.0 to 0.4}

\textit{DownStairs: friction: 1.2 to 1.5, flatness: 0.0 to 14.375, slope: -0.26 to 0.0}

\textit{Marble Slope Uphill: friction: 0.7 to 1.1, flatness: 2.25 to 2.625, slope: 0.15 to 0.25}

\textit{Marble Slope Downhill: friction: 0.7 to 1.1, flatness: 3.0 to 3.375, slope: -0.3 to -0.18}

\textit{Grassland: friction: 0.5 to 0.7, flatness: 0.25 to 9.0, slope: 0.0}

\textit{Grassland Slope Uphill: friction: 0.5 to 0.7, flatness: 0.25 to 6.125, slope: 0.06 to 0.1}

\textit{Grassland Slope Downhill: friction: 0.5 to 0.7, flatness: 0.375 to 7.75, slope: -0.25 to -0.15}

\textit{Grass and Pebble: friction: 0.05 to 0.1, flatness: 0.0 to 25.375, slope: 0.0}

\textit{Uneven Terrains: friction: 0.6 to 1.2, flatness: 0.0 to 12.75, slope: 0.0, please note Uneven terrains have no
slope and is different from UpStairs and DownStairs.}

\textit{Grass and Sand: friction: 0.3 to 0.4, flatness: 0.25 to 5.625, slope: 0.0}

\textit{For each upload image, you firstly classify it to certain type of terrain, for example, 'Ice Surface', then
infer its property based on above pre-defined range. Make sure the property you inferred falls in above range. Also
make sure you only focus on the terrains in near front of you and the terrains what make most part of vision field. }

\textbf{\textit{User}}\textit{: Now, I have given an image to you, it contains a terrain of above settings. You need to
answer: 1) for which terrain this image contains and 2) infer a concrete value (not a range) for their friction
coefficients, flatness and slope. Please format your answer like '[Grass and Sand, 0.35, 2, 0]' and don't include other
words }

\textbf{\textit{Gemini}}\textit{: (example: [Grassland, 0.55, 3, 0])}

\endgroup

\section{More experimental results}

\begin{figure}[H]
\centering
\includegraphics[width=0.9\textwidth]{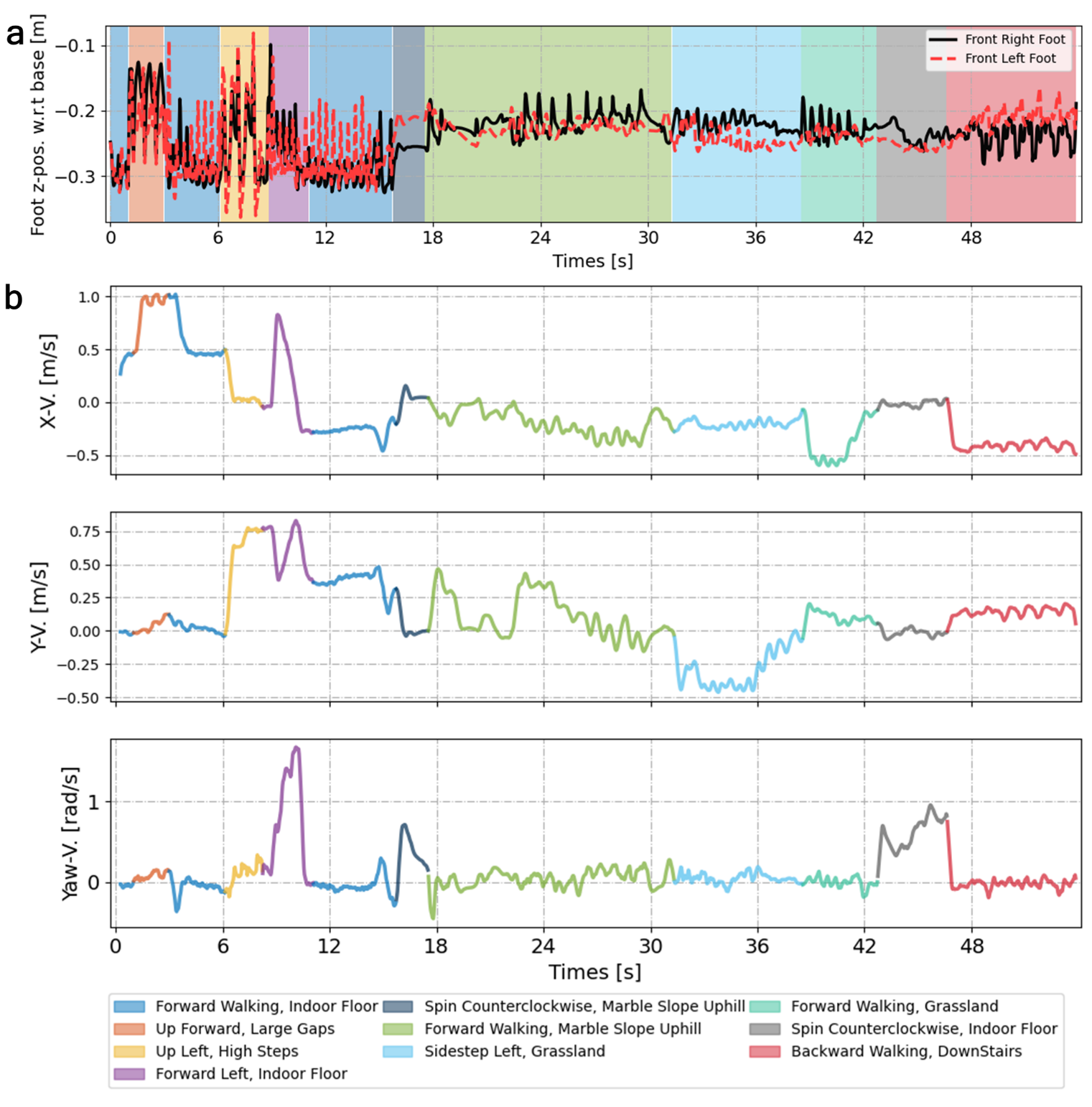}
\caption{
The profiles of two front feet positions (\textbf{a}) and the robot's body speed
(\textbf{b}) corresponding to the simulated robot parkour in Fig.~\ref{fig_The deployment of RSG for robot parkour.}\textbf{e}.
}
\label{app:fig_The profiles of two front feet positions.}
\end{figure}

\begin{figure}[H]
\centering
\includegraphics[width=0.9\textwidth]{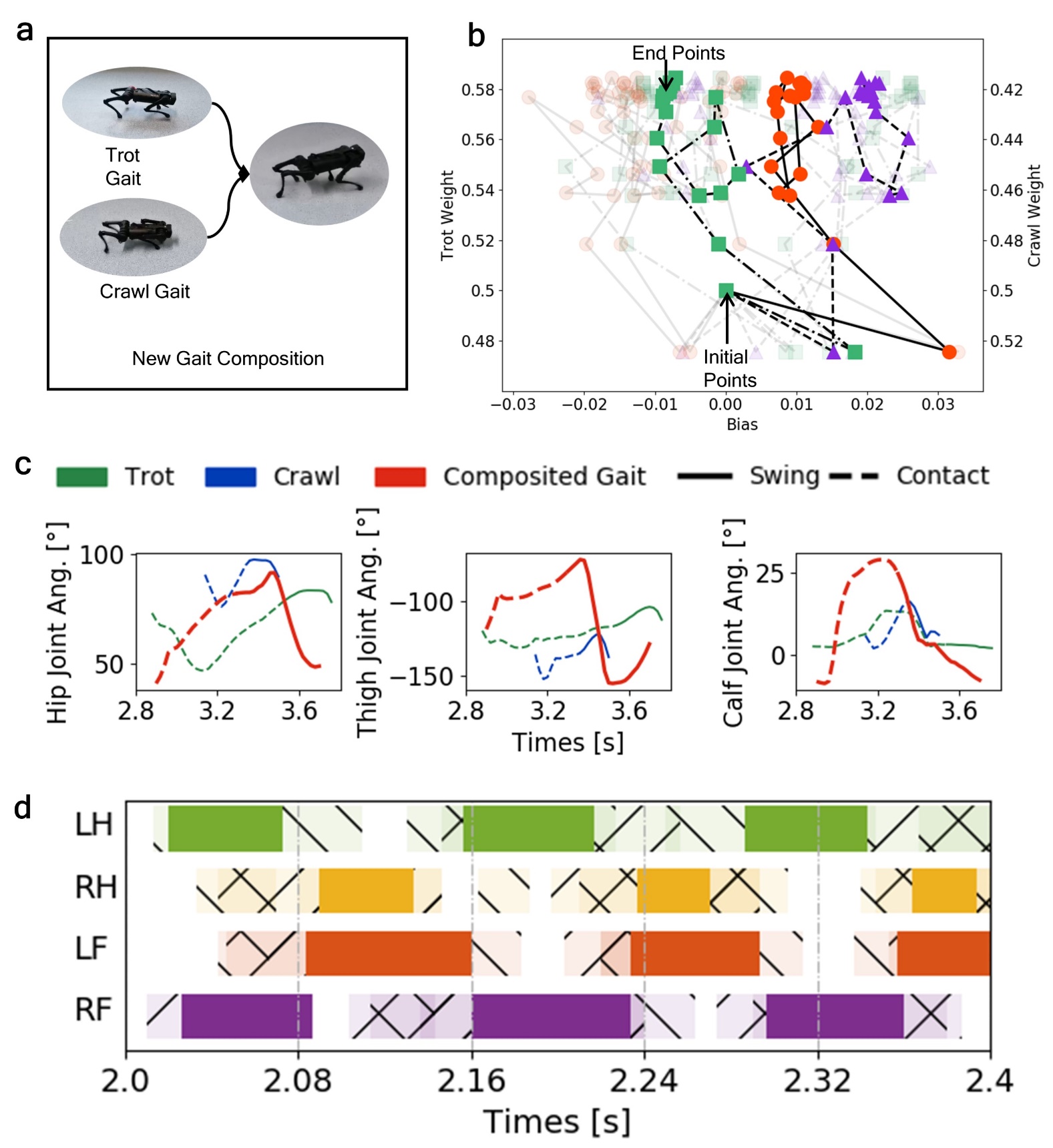}
\caption{
\textbf{a}, The novel realistic quadrupedal gait rapid learning processes. \textbf{b}, The changing
trends of biases and weights of BO corresponding to fundamental skills (left hind leg) of \textbf{a}. The shaded parts
indicate the changing trends of the other legs. \textbf{c}, The joint position of the left hind leg in one cycle
corresponding to \textbf{a}. \textbf{d}, Newly composited skill gaits corresponding to \textbf{a}. Gait patterns are
represented by the touchdown state of the robot's four legs. The shaded part indicates the gait of
the fundamental skills. $\text{\textgreek{‘}}$/' and
$\text{\textgreek{‘}}${\textbackslash}' represent \textit{trot} and \textit{crawl}. LF, left front
leg; RF, right front leg; LH, left hind leg; RH, right hind leg. 
}
\label{app:fig_The novel realistic quadrupedal gait rapid learning processes.}
\end{figure}


\newpage
\section{Tabular data and algorithms}
\begin{center}
 \scriptsize
\tablefirsthead{}
\tablehead{}
\tabletail{}
\tablelasttail{}
\tablecaption{Environment parameters.}
\label{app:Environment parameters.}
\begin{supertabular}{|m{1.1969599in}|m{1.1025599in}|m{1.1025599in}|m{1.0038599in}|}
\hline
\centering Environment Name &
\centering Friction Range &
\centering Flatness Range &
\centering\arraybslash Slope Range\\\hline
\centering Indoor Floor &
\centering [0.6, 0.9] &
\centering 0 &
\centering\arraybslash 0\\\hline
\centering Ice Surface &
\centering [0.01, 0.1] &
\centering 0 &
\centering\arraybslash 0\\\hline
\centering Upstairs &
\centering [1.2, 1.5] &
\centering [0, 13.125] &
\centering\arraybslash [0, 0.4]\\\hline
\centering Downstairs &
\centering [1.2, 1.5] &
\centering [0, 14.375] &
\centering\arraybslash [-0.26, 0]\\\hline
\centering Marble Slope Uphill &
\centering [0.7, 1.1] &
\centering [2.25, 2.625] &
\centering\arraybslash [0.15, 0.25]\\\hline
\centering Marble Slope Downhill &
\centering [0.7, 1.1] &
\centering [3.0, 3.375] &
\centering\arraybslash [-0.3, 0.18]\\\hline
\centering Grassland &
\centering [0.5, 0.7] &
\centering [0.25, 9.0] &
\centering\arraybslash 0\\\hline
\centering Grassland Slope Uphill &
\centering [0.5, 0.7] &
\centering [0.25, 6.125] &
\centering\arraybslash [0.06, 0.1]\\\hline
\centering Grassland Slope Downhill &
\centering [0.5, 0.7] &
\centering [0.375, 7.75] &
\centering\arraybslash [-0.25, 0.15]\\\hline
\centering Grass and Pebble &
\centering [0.05, 0.1] &
\centering [0, 25.375] &
\centering\arraybslash 0\\\hline
\centering Steps &
\centering [0.6, 1.2] &
\centering [0, 12.75] &
\centering\arraybslash 0\\\hline
\centering Grass and Sand &
\centering [0.3, 0.4] &
\centering [0.25, 5.625] &
\centering\arraybslash 0\\\hline
\end{supertabular}
\end{center}


\begin{center}
 \scriptsize
\tablefirsthead{}
\tablehead{}
\tabletail{}
\tablelasttail{}
\tablecaption{Name and definition of reward functions.}
\label{app:Name and definition of reward functions.}
\begin{supertabular}{|m{2.in}|m{3.2in}|}
\hline
\centering Reward Name &
\centering\arraybslash Reward Definition\\\hline
\centering Linear Velocity Tracking (LVT) &
\centering\arraybslash $ \exp \left(-\left(v_{x y}-v_{x y}^{c m d}\right)^2 / 0.25\right)$
 \\\hline
\centering Angular Velocity Tracking (AVT) &
\centering\arraybslash  $\exp \left(-\left(\omega_{y a w}-\omega_{y a v}^{c w d}\right)^2 / 0.25\right) $
 \\\hline
\centering Torques Square (TS) &
\centering\arraybslash $\tau^2$
 \ \\\hline
\centering Action Rate (AR) &
\centering\arraybslash $\left(a_{t-1}-a_t\right)^2$
 \ \\\hline
\centering Joint Acceleration (JA) &
\centering\arraybslash  $\ddot{q}^2$
 \\\hline
\centering Joint Power (JP) &
\centering\arraybslash  $|\tau \dot{q}| $
 \ \\\hline
\centering Base Height (BH) &
\centering\arraybslash $\left(h_{\text {base }}-h_{target}\right)^2$ 
 \\\hline
\centering Feet Clearance (FC) &
\centering\arraybslash $\sum_{j=0}^4\left[\left(h_{\text {base }}-p_{f, z, j}\right)-p_{f, z, j}^{\text {des }}\right]^2\left\|v_{f, x y, j}\right\|_2 $
 \ \\\hline
\centering Feet Air Time (FAT) &
\centering\arraybslash $\sum_{j=0}^4\left(t_{air, j}-0.5\right)$
 \ \\\hline
\centering Joint Power Distribution (JPD) &
\centering\arraybslash  $\operatorname{var}(\tau \dot{q})^2$
 \ \\\hline
\centering Linear Velocity Penalty (LVP) &
\centering\arraybslash  $v_z^2$
 \ \\\hline
\centering Angular Velocity Penalty (AVP) &
\centering\arraybslash  $\omega_{x y}^2$
 \ \\\hline
\centering Body Orientation Penalty (BOP) &
\centering\arraybslash   $|g|^2$
 \ \\\hline
\centering Joint Smoothness (JS) &
\centering\arraybslash  $\left(a_t-2 a_{t-1}+a_{t-2}\right)^2$
 \ \\\hline
\centering Lie Orientation (LO) &
\centering\arraybslash  $\exp \left(-\left|g_3-g_{\text {default }, 3}\right| / 0.25\right)$
 \\\hline
\centering Foot Full Contact (FFC) &
\centering\arraybslash   $\sum_{j=1}^4 \mathbf{1}_{\text {contact }, j}$
 \\\hline
\centering Jump Position (JP) &
\centering\arraybslash   $$
\left\{\begin{array}{lr}
\exp \left(-\left|h_{\text {base }}-h_{\text {target }}\right| / 0.25\right) v_z, & \text { if } v_z>0; \\
0, & \text { otherwise. }
\end{array}\right.
$$
 \\\hline
\centering Jump Body Height (JBH) &
\centering\arraybslash  $$
\begin{cases}
\exp \left(-\left|h_{\text {base }, z}-h_{\text {target }, z}\right| / 0.25\right) v_z, & \text { if } v_z>0; \\ 0, & \text { otherwise } .
\end{cases}
$$ 
 \\\hline
\centering Lie Position (LP) &
\centering\arraybslash  $$
\left\{\begin{array}{lr}
\exp \left(-\left|q-q_{\text {default }, \text { lie }}\right| / 0.25\right), & \text { if }\left|\theta_{\text {roll }}\right| \leq 0.3 ; \\
0, & \text { otherwise. }
\end{array}\right.
$$
 \\\hline
\centering Base Uprightness (BU) &
\centering\arraybslash  $1-g_3$ 
 \ \\\hline
\centering Yaw Angle Velocity (YAV) &
\centering\arraybslash  $\exp \left(-\left|\omega_{\text {yaw }}\right| / 0.25\right)$
 \\\hline
\centering Stand Hip Position (SHP) &
\centering\arraybslash  $\exp \left(-\sum_{i \in\{0,1,2,3\}}\left|q_{3 i+3}\right| / 0.25\right)$
 \\\hline
\centering Lie Orientation RPY (LORPY) &
\centering\arraybslash  $\exp \left(-\left|\theta_{\text {base }}\right| / 0.25\right)$
 \\\hline
\centering Feet Distance Penalty (FDP) &
\centering\arraybslash  $\sum_{j=0}^4\left|p_{f, y , j}\right|$
 \ \\\hline
\centering Foot Swing Height Tracking (FSHT) &
\centering\arraybslash  $\sum_j\left(p_{f, z, j}-p_{f, z, j}^{c m d}\right)^2 C_j^{c m d}\left(\theta^{c m d}, t\right)$
 \\\hline
\end{supertabular}
\end{center}


\begin{center}
 \scriptsize
\tablefirsthead{}
\tablehead{}
\tabletail{}
\tablelasttail{}
\tablecaption{Definition of tasks.}
\label{app:Definition of tasks.}
\begin{supertabular}{|m{0.8in}|m{0.9in}|m{1.6in}|m{0.5in}|m{1.2in}|}
\hline
\centering Task Name &
\centering Task Commands &
\centering Task Rewards &
\centering Use AMP? &
\centering\arraybslash Additional Information\\\hline
\centering Forward Walking &
\centering [0.6, 0, 0, 0, 0, 0] &
\centering 5LVT + 1.5AVT – 10 BH – 0.01TS – 1AR &
\centering Yes &
~
\\\hline
\centering Forward Right &
\centering [0.4, 0, 0, 0, 0, -0.4] &
\centering 5LVT + 1.5AVT – 0.01TS – 1AR &
\centering Yes &
~
\\\hline
\centering Forward Left &
\centering [0.4, 0, 0, 0, 0, 0.4] &
\centering 5LVT + 1.5AVT – 0.01TS– 1AR &
\centering Yes &
~
\\\hline
\centering Backward Walking &
\centering [-0.6, 0, 0, 0, 0, 0] &
\centering 5LVT + 1.5AVT – 0.01TS – 1AR &
\centering Yes &
~
\\\hline
\centering Backward Right &
\centering [-0.5, 0, 0, 0, 0, 0.4] &
\centering 5LVT + 1.5AVT – 0.01TS – 1AR &
\centering Yes &
~
\\\hline
\centering Backward Left &
\centering [-0.4, 0, 0, 0, 0, 0.4] &
\centering 5LVT + 1.5AVT – 0.01TS – 1AR &
\centering Yes &
~
\\\hline
\centering Sidestep Right &
\centering [0, -0.6, 0, 0, 0, 0] &
\centering 5LVT + 1.5AVT – 0.01TS – 1AR &
\centering Yes &
~
\\\hline
\centering Sidestep Left &
\centering [0, 0.6, 0, 0, 0, 0] &
\centering 5LVT + 1.5AVT – 0.01TS – 1AR &
\centering Yes &
~
\\\hline
\centering Spin Clockwise &
\centering [0, 0, 0, 0, 0, -4] &
\centering 1.5LVT + 5AVT – 0.01TS – 1AR &
\centering Yes &
~
\\\hline
\centering Spin Counter-clockwise &
\centering [0, 0, 0, 0, 0, 4] &
\centering 1.5LVT + 5AVT – 0.01TS – 1AR &
\centering Yes &
~
\\\hline
\centering Gallop &
\centering [2, 0, 0, 0, 0, 0] &
\centering 5LVT + 1.5AVT – 0.01TS – 1AR &
\centering Yes &
~
\\\hline
\centering Forward Walking Fast &
\centering [1, 0, 0, 0, 0, 0] &
\centering 5LVT + 1.5AVT – 0.01TS – 1AR &
\centering Yes &
~
\\\hline
\centering Forward Mass &
\centering [0.6, 0, 0, 0, 0, 0] &
\centering 5LVT + 1.5AVT – 0.01TS – 1AR &
\centering Yes &
\centering\arraybslash Robot base mass has twice the randomization range as other tasks\\\hline
\centering Forward Noise &
\centering [0.6, 0, 0, 0, 0, 0] &
\centering 5LVT + 1.5AVT – 0.01TS – 1AR &
\centering Yes &
\centering\arraybslash Observation noise has twice the randomization range than other tasks\\\hline
\centering Jump in Place (up) &
\centering [0, 0, 2, 0, 0, 0, 0, 0, 0.6] &
\centering 10JP – 10FFC + 10LO + 10 LORPY – 0.001TS – 0.1AR &
\centering No &
\centering\arraybslash  $h_{\text {target }}=[0,0,0.6]$
 \\\hline
\centering Jump Backward (up backward) &
\centering [-1, 0, 2, 0, 0, 0, 0, 0, 0.6] &
\centering 5LVT + 1.5AVT + 10 JBH – 10FFC + 10LO + 10LORPY – 0.001TS – 0.1AR &
\centering No &
\centering\arraybslash $h_{\text {target }, z}=0.6$
 \\\hline
\centering Jump Forward (up forward) &
\centering [1, 0, 2, 0, 0, 0, 0, 0, 0.6] &
\centering 5LVT + 1.5AVT + 10JBH – 10FFC + 10LO + 10LORPY – 0.001TS – 0.1AR &
\centering No &
\centering\arraybslash  $h_{\text {target } z}=0.6$
 \\\hline
\centering Jump Left (up left) &
\centering [0, 0.75, 2, 0, 0, 0, 0, 0, 0.6] &
\centering 5LVT + 1.5AVT + 10JBH – 10FFC + 10LO + 10LORPY – 0.001TS – 0.1AR &
\centering No &
\centering\arraybslash $h_{\text {target }, z}=0.6$
 \\\hline
\centering Jump Right (up right) &
\centering [0, -0.75, 2, 0, 0, 0, 0, 0, 0.6] &
\centering 5LVT + 1.5AVT + 10JBH – 10FFC + 10LO + 10LORPY – 0.001TS – 0.1AR &
\centering No &
\centering\arraybslash  $h_{\text {target }, z}=0.6$ 
 \\\hline
\centering Roll &
\centering [0, 0, 0, 0, 0, 0] &
\centering 3LP+1BU+1FFC+2YAV-0.000001JA-0.00001JP-0.05AR &
\centering No &
\centering\arraybslash  
 $q_{\text{default}, z }=[0 ., 1.56,-2.7, $
 
 $0., 1.56, -2.7,$ 

$0.,1.56,-2.7 $ 

$0., 1.56, -2.7]$
 \\\hline
\centering Standup (standing) &
\centering [0, 0, 0, 0, 0, 0] &
\centering 1SHP+1BU+1BH+1FFC-0.000001JA-0.00001JP-0.05AR &
\centering No &
~
\\\hline
\centering Crawl &
\centering [0.3, 0, 0, 0, 0, 0] &
\centering 1LVT+0.5AVT-1BH-1.5FC+1FAT-2LVP-0.2BOP-0.01AR-0.01AS-0.05AVP-0.00000025JA-0.00002JP-0.0000001JPD &
\centering No &
\centering\arraybslash   $h_{target}=0.15$,
$p_{f,z,j}^{des}=0.06$
 \\\hline
\centering Trot &
\centering [0.3, 0, 0, 0, 0, 0] &
\centering 1LVT+0.5AVT-1BH-0.6FSHT +1FAT-2LVP-0.2BOP-0.01AR-0.01AS-0.05AVP-0.00000025JA-0.00002JP-0.0000001JPD  &
\centering No &
\centering\arraybslash $\theta^{cmd}=(0.5,0,0)$
 \\\hline
\centering Pace &
\centering [0.3, 0, 0, 0, 0, 0] &
\centering 1LVT+0.5AVT-1BH-0.6FSHT +1FAT-2LVP-0.2BOP-0.01AR-0.01AS-0.05AVP-0.00000025JA-0.00002JP-0.0000001JPD &
\centering No &
\centering\arraybslash  $\theta^{cmd}=(0.5,0,0)$
 \\\hline
\centering Small Steps &
\centering [0.3, 0, 0, 0, 0, 0] &
\centering 1LVT+0.5AVT-1BH-1.5FC+1FAT-2LVP-0.2BOP-0.01AR-0.01AS-0.05AVP-0.00000025JA-0.00002JP-0.0000001JPD-0.5FDP &
\centering No &
~
\\\hline
\centering Others &
\centering Backward Walking Slow: [-0.3, 0, 0, 0, 0, 0]; Forward Walking Slow: [0.3, 0, 0, 0, 0, 0]; Sidestep Left Slow:
[0, 0.3, 0, 0, 0, 0]; idestep Right Slow: [0.,0.3, 0, 0, 0, 0, 0]; Spin Clockwise Slow: [0, 0, 0, 0, 0, -0.8]; Spin
Counterclockwise Slow: [0., 0, 0, 0, 0, 0.8] &
\centering 1LVT+0.5AVT-1BH-1.5FC+1FAT-2LVP-0.2BOP-0.01AR-0.01AS-0.05AVP-0.00000025JA-0.00002JP-0.0000001JPD &
\centering No &
\centering\arraybslash 
 $h_{target}=0.25$,
$p_{f,z,j}^{des}=0.26$
\\\hline
\end{supertabular}
\end{center}


\begin{center}
 \scriptsize
\tablefirsthead{}
\tablehead{}
\tabletail{}
\tablelasttail{}
\tablecaption{Definition of robot parameters and domain randomization.}
\label{app:Definition of robot parameters and domain randomization.}
\begin{supertabular}{|m{1.8900598in}|m{0.70875984in}|m{1.1191599in}|m{0.7261598in}|}
\hline
\centering Parameter (Dimension) &
\centering Unit &
\centering Randomization/Noise Range &
\centering\arraybslash Operator\\\hline
\centering Projected Gravity (3) &
\centering {}- &
\centering [-0.05, 0.05] &
\centering\arraybslash additive\\\hline
\centering Joint Position (12) &
\centering rad &
\centering [-0.03, 0.03] &
\centering\arraybslash additive\\\hline
\centering Joint Velocity (12) &
\centering rad/s &
\centering [-1.5, 1.5] &
\centering\arraybslash additive\\\hline
\centering Base Angular Velocity (3) &
\centering rad/s &
\centering [-0.3, 0.3] &
\centering\arraybslash additive\\\hline
\centering Base x-y Linear Velocity (2) &
\centering m/s &
\centering [-0.1, 0.1] &
\centering\arraybslash additive\\\hline
\centering Base Mass (1) &
\centering kg &
\centering [-1, 1] &
\centering\arraybslash additive\\\hline
\centering Kp Factor (12) &
\centering \% &
\centering [0.9, 1.1] &
\centering\arraybslash scaling\\\hline
\centering Kd Factor (12) &
\centering \% &
\centering [0.9, 1.1] &
\centering\arraybslash scaling\\\hline
\end{supertabular}
\end{center}


\begin{center}
\scriptsize
\tablefirsthead{}
\tablehead{}
\tabletail{}
\tablelasttail{}
\tablecaption{Hyperparameters of PPO algorithm.}
\label{app:Hyperparameters of PPO algorithm.}
\begin{supertabular}{|m{2.44486in}|m{1.3309599in}|}
\hline
\centering Hyperparameter &
\centering\arraybslash Value\\\hline
\centering Actor hidden size &
\centering\arraybslash [512, 256, 128]\\\hline
\centering Critic hidden size &
\centering\arraybslash [512, 256, 128]\\\hline
\centering CEN encoder hidden size &
\centering\arraybslash [128, 64]\\\hline
\centering CEN decoder hidden size &
\centering\arraybslash [64, 128]\\\hline
\centering CEN hidden state dim &
\centering\arraybslash 16\\\hline
\centering AMP discriminator hidden size &
\centering\arraybslash [1024, 512]\\\hline
\centering Learning rate &
\centering\arraybslash Adaptive\\\hline
\centering Optimizer &
\centering\arraybslash Adam\\\hline
\centering Value loss clipping &
\centering\arraybslash True\\\hline
\centering Learning epochs per iteration &
\centering\arraybslash 5\\\hline
\centering Minibatch size &
\centering\arraybslash 6144\\\hline
\centering Discount factor &
\centering\arraybslash 0.99\\\hline
\centering GAE factor &
\centering\arraybslash 0.95\\\hline
\centering Maximum gradient norm &
\centering\arraybslash 1\\\hline
\end{supertabular}
\end{center}

\begin{center}
\scriptsize
\tablefirsthead{}
\tablehead{}
\tabletail{}
\tablelasttail{}
\tablecaption{Hyperparameters of the RSG inference and composition.}
\label{app:Hyperparameters of the RSG inference and composition.}
\begin{supertabular}{|m{2.87126in}|m{0.6094598in}|}
\hline
\centering Hyperparameter &
\centering\arraybslash Value\\\hline
\centering High match threshold $\alpha_{high}$
  &
\centering\arraybslash 0.9\\\hline
\centering Low match threshold $\alpha_{low}$
  &
\centering\arraybslash 0.7\\\hline
\centering Mean function  $m_{0}$
  &
\centering\arraybslash 0\\\hline
\centering Parameters of the kernel function $\sigma_{f},l$
  &
\centering\arraybslash 2, 1\\\hline
\centering Measurement noise standard deviation $\sigma_{noise}$
  &
\centering\arraybslash $10^{-4}$\\\hline
\end{supertabular}
\end{center}

\begin{algorithm}
\caption{RSG construction and representation.}
\begin{algorithmic}[1]
\State \textbf{Input:} Dataset  $D$
 of fact triplets  $\left(c, r_{c \rightarrow s}, s\right)$
 , environment encoder $\phi_e$
 , task encoder  $\phi_t$
 , all skills  $\{s\}$
 , environment to skill  $r_{e \rightarrow s}$ and task to skill  $r_{t \rightarrow s}$.
\State \textbf{Output:} trained  $\phi_e, \phi_t,\{s\}, r_{e \rightarrow s}$ and $r_{t \rightarrow s}$.
\State Sample mini-batch   $B$ from  $D$.
\State Augment $B$ with negative and soft triplets.
\State Encoder $c$
 with corresponding $\phi_c$.
 \State Compute loss according to $$
\left.\mathcal{L}=\left(\mathbb{S}_{positive }-1\right)^2+\left(\mathbb{S}_{negative }-0\right)^2+\max \left(0, \quad \mathbb{S}_{soft }-1+\delta\right)\right),
$$
where $\mathbb{S}=e^{-\lambda \|\left(c-w_r^T c w_r\right)+d_r-\left(s-w_r^T s w_r\right) \|}$.
\State Backpropagate the loss and perform stochastic gradient descent to train until convergence.
\end{algorithmic}
\label{app:RSG construction and representation.}
\end{algorithm}





\begin{algorithm}
\caption{RSG inference and usage.}
\begin{algorithmic}[1]
\State \textbf{Initialization:} Given new env. and task query  $q_{\text {env }}, q_{\text {task }}$; 
The constructed RSG   $\Omega$; 
The skill inference score threshold $\alpha_{\text {high }}, \alpha_{\text {low }}$; 
The skill inference score function $\mathbb{S}(\cdot)$.
\State {Calculate the skill inference scores corresponding to the new skills required for the
downstream task $p=\mathbb{S}\left(q_{env.}, q_{\text {task }}\right)$
 .}
 \State Return the inferred skills $\left[s_0,\cdots,s_N\right]=\Omega(p)$.
 \State If $\alpha_{\text {high }}=0.9 \leq \mathbb{S} \leq 1$ 
 : Directly deploy the first inferred skill. 
 \State Else If $\alpha_{\text {low }}=0.7 \leq \mathbb{S}<\alpha_{\text {high }}=0.9$
 : The inferred skills are optimized in action space by BO.
 \State Else If $0 \leq \mathbb{S}<\alpha_{\text {lov }}=0.7$
 : Perform online RL fine-tuning on the inferred skills.
\end{algorithmic}
\label{app:RSG inference and usage.}
\end{algorithm}



\bibliographystyle{elsarticle-num} 
\bibliography{elsarticle-template-num}







\end{document}